\documentclass{article}

\PassOptionsToPackage{numbers}{natbib}
\PassOptionsToPackage{table,dvipsnames}{xcolor}

\usepackage[preprint]{neurips_2025}

\usepackage[utf8]{inputenc}
\usepackage[T1]{fontenc}

\usepackage{amsmath,amsfonts}
\usepackage{nicefrac,booktabs,microtype,url,graphicx}
\usepackage{xcolor}              

\usepackage{algorithm}
\usepackage[noend]{algpseudocode}

\usepackage{multirow,makecell,enumitem}

\usepackage{caption,subcaption,wrapfig}

\usepackage{tcolorbox,lipsum,soul,pifont,appendix,tocloft,
            circledsteps,fontawesome,marvosym,float}

\usepackage{natbib}
\usepackage[
  pagebackref=true,
  breaklinks=true,
  colorlinks=true,
  citecolor=blue,
  linkcolor=red,
  bookmarks=false
]{hyperref}

\usepackage{amssymb}
\usepackage{array}

\definecolor{lightgray}{gray}{0.75}  

\definecolor{blue}{rgb}{0.21,0.49,0.74}
\definecolor{red}{rgb}{0.8, 0.2, 0.2}
\definecolor{green}{rgb}{0, 0.5, 0}
\definecolor{yellow}{RGB}{218, 160, 109}

\title{AirV2X: Unified Air-Ground\\Vehicle-to-Everything Collaboration}

\author{
    \bf  Xiangbo Gao$^1$, Yuheng Wu$^2$, Fengze Yang$^3$, Xuewen Luo$^3$, Keshu Wu$^1$, Xinghao Chen$^4$, \\
    \bf Yuping Wang$^5$, Chenxi Liu$^3$, Yang Zhou$^1$, Zhengzhong Tu$^{1}$\thanks{Corresponding Author: Zhengzhong Tu (\texttt{tzz@tamu.edu})}\\[2pt]
  $^1$Texas A\&M University, 
  $^2$KAIST, 
  $^3$The University of Utah,
  $^4$University of Washington, \\
  $^5$University of Michigan
}

\begin{document}

\maketitle

\begin{abstract}

While multi-vehicular collaborative driving demonstrates clear advantages over single-vehicle autonomy, traditional infrastructure-based V2X systems remain constrained by substantial deployment costs and the creation of "uncovered danger zones" in rural and suburban areas. We present AirV2X-Perception, a large-scale dataset that leverages Unmanned Aerial Vehicles (UAVs) as a flexible alternative or complement to fixed Road-Side Units (RSUs). Drones offer unique advantages over ground-based perception: complementary bird's-eye-views that reduce occlusions, dynamic positioning capabilities that enable hovering, patrolling, and escorting navigation rules, and significantly lower deployment costs compared to fixed infrastructure. Our dataset comprises 6.73 hours of drone-assisted driving scenarios across urban, suburban, and rural environments with varied weather and lighting conditions. The AirV2X-Perception dataset facilitates the development and standardized evaluation of Vehicle-to-Drone (V2D) algorithms, addressing a critical gap in the rapidly expanding field of aerial-assisted autonomous driving systems. The dataset and development kits are open-sourced at \href{https://github.com/taco-group/AirV2X-Perception}{https://github.com/taco-group/AirV2X-Perception}.

\end{abstract}

\section{Introduction}

Multi-vehicular collaborative driving has been shown to be more effective than single-vehicle autonomous driving primarily due to the complementary sensory coverage provided through inter-vehicle cooperation. This collaborative approach significantly reduces perception limitations~\cite{zhang2023perception, wang2025generative, gao2024mambast} by enabling vehicles to share sensor data and compensate for each other's blind spots. Recent research in Vehicle-to-Everything (V2X) communication demonstrates that Road-Side Units (RSUs) mounted on infrastructure offer superior perception capabilities than vehicle-only collaboration, primarily due to their elevated positioning that minimizes blind spots and provides broader field-of-view coverage. However, traditional infrastructure-based solutions face significant economic constraints. The substantial construction and maintenance costs associated with RSUs necessitate strategic deployment decisions. Consequently, these units are predominantly installed at high-traffic intersections and critical urban junctions to maximize cost-benefit ratios. This economically-driven compromise results in what we term ``uncovered danger zones''---highways, pedestrian zones, suburban neighborhoods, and rural areas where vehicles operate without the perceptual advantages of infrastructure support.

\paragraph{Motivation.} In parallel with these infrastructure limitations, the rapid rise of the low‑altitude economy and steady advances in Unmanned Aerial Vehicle (UAV) technology offer a promising alternative. Modern drones now handle everything from emergency response and fire rescue to goods transport and everyday food delivery. The evolution of drone capabilities has naturally paved the way for drone-based perception systems, which represent a logical extension of this technology into the autonomous driving domain. Unlike static infrastructure-dependent solutions, drone-assisted collaborative driving leverages the mobility and versatility of aerial platforms to enhance vehicular perception in environments where traditional RSUs are economically unfeasible. Compared to traditional RSUs, drone-based perception systems offer several significant advantages:

\begin{itemize}[leftmargin=8pt, itemsep=2pt, parsep=2pt, topsep=0pt, partopsep=0pt]
\item \textbf{Drones offer a unique and complementary perception perspective} distinct from both vehicle-mounted sensors and fixed RSUs. Ground vehicles capture primarily horizontal, ego-centric views limited by occlusions, while RSUs provide fixed vantage points typically restricted to intersections. In contrast, drones deliver real-time bird's-eye views that are both elevated and mobile, enabling holistic scene understanding in cluttered environments. This aerial perspective facilitates tracking of agents across occlusions, better anticipation of multi-agent interactions, and improved detection of lane changes or cut-ins otherwise hidden from ground-level sensors.
\item \textbf{Drones provide superior operational adaptability and flexibility} through dynamic positioning based on real-time requirements. They can \underline{hover} above specific areas of interest, \underline{patrol} designated routes, or be programmed to \underline{escort} individual vehicles, delivering tailored perception assistance precisely where and when needed—capabilities that fixed infrastructure simply cannot match.
\item \textbf{Drone solutions present a more cost-effective approach to expanding coverage}. According to the 5G Automotive Association (5GAA) in 2020~\cite{nokes2020v2i}, a fixed RSU without camera or lidar sensors costs between $20,000-$80,000. In comparison, fully-equipped drones with high-resolution cameras and lidar range from under $1,000$ to $50,000$ depending on specifications. This economic flexibility enables strategic resource allocation—deploying less expensive Vehicle-to-Drone (V2D) systems in rural areas while positioning advanced systems in accident-prone or high-density zones.
\item \textbf{V2D systems can be integrated with emerging aerial networks} as the low-altitude economy develops. The increasing presence of UAVs performing various tasks creates opportunities for "opportunistic" sensing and communication channels for ground vehicles. These aerial vehicles, already equipped with advanced sensors and communication devices, can establish a multi-layered perception network with minimal additional infrastructure investment, creating a more robust and redundant system for autonomous vehicle perception.
\end{itemize}

These compelling advantages have sparked considerable interest in Vehicle-to-Drone (V2D) communication systems, leading to substantial research efforts focused on developing algorithms that can effectively utilize aerial perspectives to enhance ground vehicle autonomous driving. As a result, there has been a significant increase in research exploring various aspects of drone-assisted vehicular perception, from communication protocols to collaborative sensing frameworks~\cite{hu2022where2comm,gao2025stamp,gao2025langcoop,wu2025synthetic}. However, with the flourishing of V2D algorithms and the demand for more advanced solutions, there remains a critical absence of high-quality datasets specifically designed for training and evaluating drone-assisted perception systems. This limitation hampers the development of robust algorithms that can effectively leverage aerial perspectives to enhance vehicular perception.

To bridge this gap, we present \textbf{AirV2X-Perception}, \textbf{a large-scale drone-assisted V2X driving dataset}. AirV2X-Perception contains 6.73 hours of drone-assisted driving data collected in the co-simulation of the CARLA ~\cite{dosovitskiy2017carla} and Airsim~\cite{shah2018airsim} simulator. The dataset contains multiple connected agent types including vehicles, roadside units (RSU), and drones, each equipped with different sensors. The dataset is collected in various urban and rural areas, with different weather (clear, rainy, foggy, and cloudy) and lighting conditions (daytime, dusk, and nighttime). AirV2X-Perception also contains various common navigation strategies for drones, including hovering, patrolling, and escorting, to provide a more comprehensive evaluation of the V2D algorithms. To accommodate the challenges of large-scale perception networks, we include up to 15 connected agents simultaneously (5 vehicles, 5 RSUs, 5 drones) in single scenarios. By releasing this comprehensive dataset to the research community, we aim to accelerate the development of robust drone-assisted perception algorithms and establish standardized benchmarks for performance evaluation.

\begin{table}[htbp]
  \centering
  \small
  \setlength{\tabcolsep}{6.3pt}
\caption{`--' denotes an unavailable attribute. `R/S' refers to whether the data is real‑world or simulated. `A.W.' denotes adverse weather conditions. `D/N' indicates daytime or nighttime scenarios, while `U/R' refers to urban or rural environments.}
  \begin{tabular}{@{} l l l c c c c c c c c @{}}
    \toprule
    \multicolumn{3}{c}{} &
    \multicolumn{3}{c}{\bf Max Connected Agents} &
    \multicolumn{2}{c}{\bf Modalities} &
    \multicolumn{3}{c}{\bf Diversity} \\
    \cmidrule(lr){4-6}\cmidrule(lr){7-8}\cmidrule(lr){9-11}
    \bf Year & \bf Dataset & \bf R/S &
    \bf CAVs & \bf Infra & \bf Drones &
    \bf Cam. & \bf LiDAR &
    \bf A.W. & \bf D/N & \bf U/R \\
    \midrule
    \multicolumn{11}{c}{\it Vehicle + Infrastructure}\\
    \hline
    2021 & OPV2V~\cite{xu2022opv2v}     & Sim  & 7  & \textcolor{lightgray}{0} & \textcolor{lightgray}{0} & $\checkmark$ & $\checkmark$ & \textcolor{lightgray}{--}            & D              & U \\
    2022 & V2XSim~\cite{li2022v2x}      & Sim  & 5  & 1  & \textcolor{lightgray}{0} & $\checkmark$ & $\checkmark$ & \textcolor{lightgray}{--}            & D            & U \\
    2022 & V2XSet~\cite{xu2022v2xvit}   & Sim  & 7  & 1  & \textcolor{lightgray}{0} & $\checkmark$ & $\checkmark$ & \textcolor{lightgray}{--}            & D            & U \\
    2022 & DAIR-V2X~\cite{yu2022dair}   & Real & 2  & 1  & \textcolor{lightgray}{0} & $\checkmark$ & $\checkmark$ & \textcolor{lightgray}{--}            & D+N  & U \\
    2023 & V2V4Real~\cite{xu2023v2v4real} & Real & 2  & \textcolor{lightgray}{0} & \textcolor{lightgray}{0} &       \textcolor{lightgray}{--}       & $\checkmark$ & \textcolor{lightgray}{--}            & D            & U \\
    2024 & Rcooper~\cite{hao2024rcooper}            & Real & \textcolor{lightgray}{0} & 4  & \textcolor{lightgray}{0} & $\checkmark$ & $\checkmark$ & \textcolor{lightgray}{--}            & D+N  & U \\
    2024 & TUMTraf~\cite{zimmer2024tumtraf}        & Real & 1  & 1  & \textcolor{lightgray}{0} & $\checkmark$ & $\checkmark$ & $\checkmark$  & D            & U \\
    2024 & HoloVIC~\cite{ma2024holovic}            & Real & 1  & 4  & \textcolor{lightgray}{0} & $\checkmark$ & $\checkmark$ & \textcolor{lightgray}{--}            & D            & U \\
    2024 & V2XReal~\cite{xiang2024v2xreal}            & Real & 4  & 2  & \textcolor{lightgray}{0} & $\checkmark$ & $\checkmark$ & \textcolor{lightgray}{--}            & D            & U \\
    2025 & V2X-Radar~\cite{huang2024v2xr}          & Real & 1  & 1  & \textcolor{lightgray}{0} & $\checkmark$ & $\checkmark$ & $\checkmark$  & D+N  & U \\
    \midrule
    \multicolumn{11}{c}{\it Drones Only}\\
    \hline
    2022 & CoP-UAV~\cite{hu2022where2comm}   & Sim  & \textcolor{lightgray}{0} & \textcolor{lightgray}{0} & 5  & $\checkmark$ & \textcolor{lightgray}{--}           & \textcolor{lightgray}{--}            & D            & U \\
    2023 & CoP-UAV+~\cite{hu2023collaboration}  & Sim  & \textcolor{lightgray}{0} & \textcolor{lightgray}{0} & 10 & $\checkmark$ & \textcolor{lightgray}{--} & \textcolor{lightgray}{--}            & D            & U \\
    2024 & UAV3D~\cite{ye2024uav3d}              & Sim  & \textcolor{lightgray}{0} & \textcolor{lightgray}{0} & 5  & $\checkmark$ & \textcolor{lightgray}{--}           & \textcolor{lightgray}{--}            & D            & U \\
    2024 & U2Udata~\cite{feng2024u2udata}            & Sim  & \textcolor{lightgray}{0} & \textcolor{lightgray}{0} & 3  & $\checkmark$ & $\checkmark$ & $\checkmark$  & D+N  & R \\
    \midrule
    \multicolumn{11}{c}{ \it Vehicle + Infrastructure + Drones}\\
    \hline
    2023 & MAVREC~\cite{dutta2024mavrec}             & Real & 1  & \textcolor{lightgray}{0} & 1  & $\checkmark$ & \textcolor{lightgray}{--}           & \textcolor{lightgray}{--}            & D            & U+R \\
    2024 & UVCPNet~\cite{wang2024uvcpnet}            & Sim  & 1  & \textcolor{lightgray}{0} & 2  & $\checkmark$ & \textcolor{lightgray}{--}           & \textcolor{lightgray}{--}            & D            & U \\
    2025 & Griffin~\cite{wang2025griffin} & Sim  & 1 & 0  & 1 & $\checkmark$ & $\checkmark$ & $\checkmark$  & D+N  & U+R \\
    2025 & AGC Drive~\cite{hou2025agc} & Real  & 2 & 0  & 1 & $\checkmark$ & $\checkmark$ & \textcolor{lightgray}{--}  & D+N  & U+R \\
    \rowcolor{pink!40}
    2025 & AirV2X (Ours) & Sim  & 5 & 5  & 5 & $\checkmark$ & $\checkmark$ & $\checkmark$  & D+N  & U+R \\
    \bottomrule
  \end{tabular}
  \label{tab:v2x-datasets}
  \vspace{-5mm}
\end{table}

\section{Related Works}

\subsection{V2X Datasets}
\label{sec:related_works_v2x_datasets}

 In this section, we summarize some widely used V2X datasets in Table~\ref{tab:v2x-datasets}. Existing V2X datasets~\cite{wang2025collaborative} can be categorized into three distinct groups. The \textbf{Vehicle + Infrastructure} category constitutes the majority of current datasets, with simulated environments like OPV2V~\cite{xu2022opv2v}, V2XSim~\cite{li2022v2x}, and V2XSet~\cite{xu2022v2xvit} providing multiple connected vehicles and infrastructure elements for various perception tasks. Real-world counterparts such as DAIR-V2X~\cite{yu2022dair}, TUMTraf~\cite{zimmer2024tumtraf}, and V2XReal~\cite{xiang2024v2xreal} offer authentic data but typically suffer from limited scale and environmental diversity, predominantly focusing on urban daytime scenarios. The \textbf{Drone-specific} category includes datasets such as CoPercpetion-UAV~\cite{hu2022where2comm}, CoPercpetion-UAV+~\cite{hu2023collaboration}, UAV3D~\cite{ye2024uav3d}, and U2Udata~\cite{feng2024u2udata}, which concentrate on aerial vehicle collaboration but lack the ground-vehicle components essential for comprehensive V2X research. The emerging \textbf{Vehicle + Infrastructure + Drone} category attempts to integrate all three agent types, but current offerings like MAVREC~\cite{dutta2024mavrec}, UVCPNet~\cite{wang2024uvcpnet},  Griffin~\cite{wang2025griffin}, and AGC Drive~\cite{hou2025agc} exhibit significant limitations—restricted to minimal agent configurations (typically one vehicle with one or two drones), supporting limited perception tasks, and lacking environmental diversity. Our proposed AirV2X-Perception dataset addresses these limitations by providing an unprecedented comprehensive solution integrating all three agent types at scale. With support for 5 CAVs, 5 infrastructure elements, and 5 drones, it offers the most extensive connected agent environment currently available. Unlike existing datasets with limited scenarios, AirV2X-Perception encompasses diverse environmental conditions spanning urban and rural settings, daytime and nighttime operations, and various adverse weather conditions. Furthermore, it supports both camera and LiDAR modalities across vehicular and RSU agent types alongside high-resolution camera sensors for drone agents, enabling research on multi-modal and cross-modal perception algorithms. The dataset facilitates various perception tasks including object detection, semantic segmentation, and tracking, making it versatile for different collaborative perception research directions while providing a realistic testbed for evaluating algorithms under conditions closer to real-world deployment scenarios.

\subsection{V2X Perception Algorithms}

\textbf{Fusion scheme taxonomy} is a primary categorization framework for V2X collaborative perception algorithms. \textbf{Early fusion} \citep{gao2018object, chen2019cooper, arnold2020cooperative} involve direct sharing of raw sensor data, maximally preserving information but requiring prohibitive bandwidth for practical deployment. \textbf{Late fusion} methods \citep{melotti2020multimodal, fu2020depth, zeng2020dsdnet, shi2022vips, glaser2023we} share only final predictions, dramatically reducing communication overhead at the cost of suboptimal accuracy due to information loss. \textbf{Intermediate fusion} techniques \citep{wang2020v2vnet, liu2020when2com, cui2022coopernaut, xu2022v2x, qiao2023adaptive, li2023learning, wang2023core, yu2023vehicle, wang2025cocmt} represent the most widely adopted approach, striking a balance by sharing mid-level representations (e.g., BEV features) that enable flexible collaboration while maintaining reasonable data transmission bandwidth. The emerging \textbf{language-based fusion} paradigm~\citep{luo2025senserag, you2024v2x, wu2025v2x, gao2025langcoop, luo2025v2x} offers advantages in transmission efficiency, explainability, and interoperability, though our work focuses primarily on benchmarking the predominant intermediate fusion approaches.

\textbf{Key technical challenges} in V2X collaborative perception include model complexity, communication efficiency, and agent heterogeneity. Early transformer-based architectures like V2X-ViT \cite{xu2022v2xvit} and CoBEVT \cite{xu2022cobevt} achieved significant performance improvements over predecessors such as F-Cooper \cite{chen2019f} and V2VNet \cite{wang2020v2vnet}, but at the cost of \textbf{high computational demands}. Subsequent work has addressed these limitations through various strategies: SICP~\cite{qu2024sicp} employs convolutional neural networks to reduce model complexity, while HEAL \cite{lu2024extensible} and STAMP \cite{gao2025stamp} implement hierarchical fusion strategies that enhance feature processing while scaling to larger agent numbers. For \textbf{communication efficiency}, When2com \cite{liu2020when2com} introduced selective communication through graph grouping, and Where2comm~\cite{hu2022where2comm} utilized spatial confidence maps to share only partial feature maps—significantly reducing bandwidth requirements while preserving accuracy. \textbf{Agent heterogeneity} presents another challenge in V2X systems. HEAL \cite{lu2024extensible} addressed this through backward alignment strategies. STAMP \cite{gao2025stamp} developed lightweight adapter-reverter pairs for feature alignment without modifying the local models. LangCoop \cite{gao2025langcoop} innovated using natural language as a universal communication medium between diverse agents. Our AirV2X-Perception dataset features 10+ connected agents per scene and introduces the novel heterogeneity of aerial perspectives. It serves as a comprehensive benchmark for evaluating these algorithms against multiple real-world challenges, particularly in the underexplored domain of drone-assisted V2X systems.

\section{AirV2X-Perception Dataset}

\label{sec:dataset}

In this section, we introduce AirV2X-Perception, a novel dataset designed specifically to advance drone-assisted V2X collaborative perception research. We first detail our simulation environment and scenario design (\S \ref{sec:environment}), followed by an exploration of three distinct drone navigation strategies—hover, patrol, and escort (\S \ref{sec:trajectory}). We then describe our data collection methodology (\S \ref{sec:data_collection}) and conclude with our annotation approach and downstream task formulations (\S \ref{sec:annotation}), designed to facilitate benchmark evaluations for this emerging research domain.

\subsection{Simulator Environment and Scenario Design}

\label{sec:environment}

\noindent \textbf{Simulator Environment:} The AirV2X-Perception dataset is collected by co-simulating CARLA~\cite{dosovitskiy2017carla} and Airsim~\cite{shah2018airsim} simulator environments. CARLA simulator is a high-fidelity open-source simulator for autonomous driving research, which provides realistic physical dynamics, vehicle control and interactions as well as photo-realistic digital assets for sensor data collection. Airsim is designed for unmanned aerial vehicle (UAV) simulation and provides realistic physical dynamics of drones. In this project, we use it to simulate the dynamics of drone agents and synchronized with actors in the CARLA simulator. We designed a total of 6.73-hour V2X collaborative driving sequences collected through Towns 1-4 and 6-7, and 12 of the CARLA simulator. We ignore Towns 5 and 10 since these maps contain static objects embedded into the CARLA static map asset that do not have accurate ground truth labels or sensing results, which may potentially lead to unstable model training. We also ignore town 11 as it is an unadorned map that does not align with the collaborative perception objectives of this work. 

\begin{figure}[!ht]
    \centering
    \includegraphics[width=\linewidth]{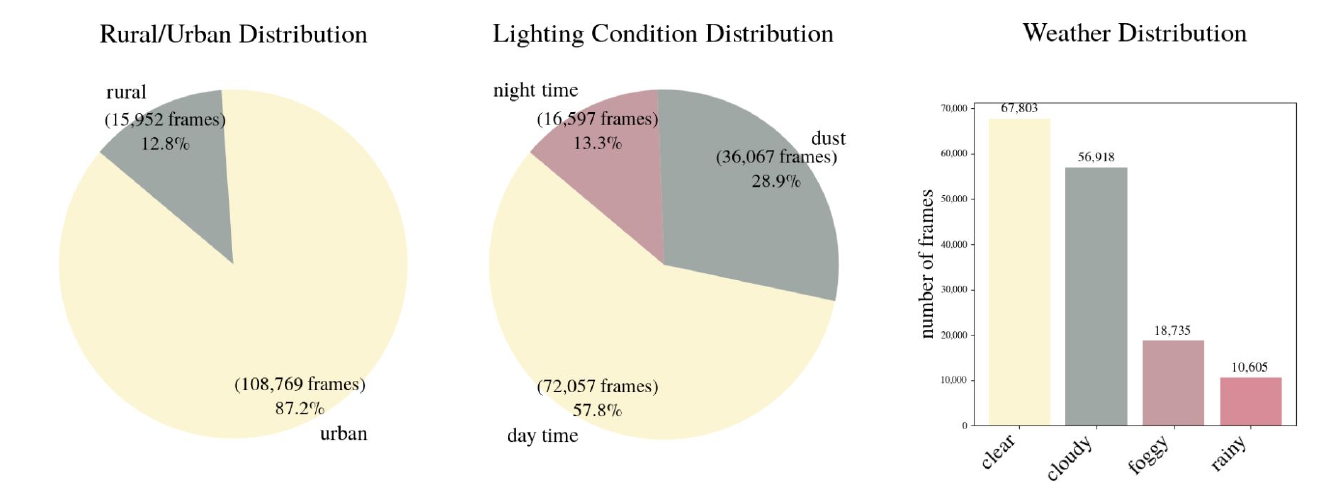}
    \caption{The percentage of each scenario in the dataset including environments, lighting conditions, and weathers.}
    \label{fig:scene}
\end{figure}

\noindent \textbf{Scenario Design:} The dataset is collected in four different weather conditions including clear, cloudy, foggy, and rainy, as well as three daytime type variations including day, dusk, and night, forming a variety of lighting conditions such as cloudy day, rainy night, and clear dusk, etc., assisting model training and evaluation in different weather and lighting conditions. Since the CARLA simulator provides urban and rural maps, we also collect data in both environments. The statistical distribution of the scenes in the dataset is shown in Figure~\ref{fig:scene}.

\subsection{Navigation Strategies for Drones}
\label{sec:trajectory}

\begin{figure}[!ht]
    \centering
    \includegraphics[width=\linewidth]{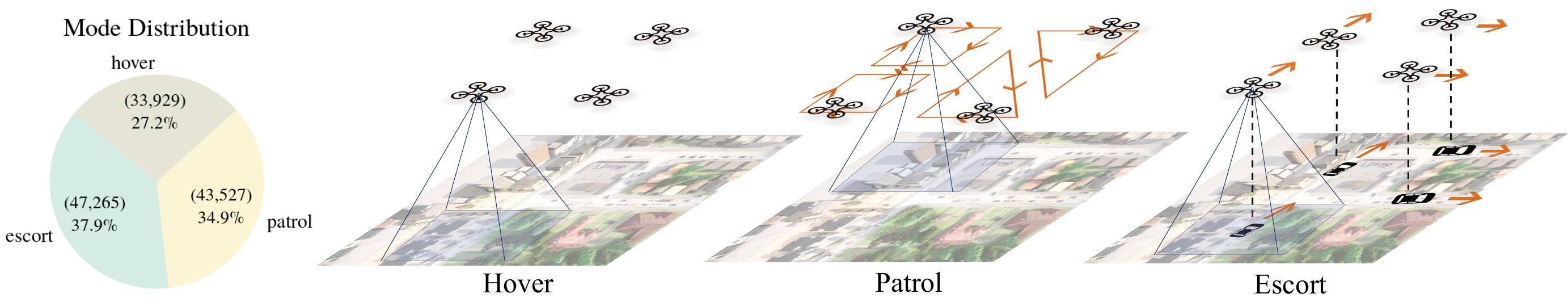}
    \caption{A conceptual visualization of three different navigation strategies for drones, along with the data distribution of each strategy.}
    \label{fig:planning}
\end{figure}

We designed three types of navigation strategies for drones, hover -- letting a drone hover at a fixed position, patrol -- assigning each drone a list of predefined waypoints to navigate, and escort -- assigning each drone a connected vehicle to follow. Each of these navigation strategy designs has its own advantages and application in the real world. More specifically, hover is suitable for single-point monitoring in a specific area with minimum power consumption. Secondly, patrol enables the drone to cover a larger area, which is more suitable for large and complex transportation scenarios, but requires more power consumption and high complexity in multi-drone route planning. Finally, escort strategy is designed to assist a certain vehicle or a certain platooning group of vehicles. A conceptual visualization of these three trajectories is shown in Figure \ref{fig:planning}. The dataset incorporates all of these three types of trajectories. The distribution of each type of navigation strategy in the dataset is shown in Figure~\ref{fig:planning}.

\subsection{Data Collection}
\label{sec:data_collection}

Sensor data are collected at a 5 Hz frequency. Each driving sequence contains 3 to 5 connected vehicles, 3 to 5 connected drones, and 3 to 5 RSUs. Each agent is equipped with both LiDAR and camera sensors and the details of the sensor setups are summarized in Table~\ref{tab:sensor_setup}. Each vehicle is equipped with 6 cameras forming a surround-view matrix. The camera sensor placement can be visualized in Figure \ref{fig:anno} (c). Each agent also carries GNSS sensors for global positioning while vehicles and drones are also equipped with IMU sensors. LiDAR point clouds are visualized in Figure \ref{fig:anno} (b). Sensors from vehicles, drones, and RSUs are annotated with different colors. LiDAR point clouds of different types of agents complement each other, forming more complete point clouds of the scene.

\begin{table}[!ht]
  \centering
  \small
  \caption{Sensor configurations for vehicles, RSUs, and drones.}
  \label{tab:sensor_setup}
  \begin{tabular}{@{} l p{4.5cm} p{5cm} l @{}}
    \toprule
    \bf Agent & \bf  LiDAR & \bf  Cameras & \bf  Other Sensors \\
    \midrule
    Vehicle & 
      1$\times$ 360° 64-channel LiDAR; 20 Hz rotation; Vertical FOV +10°/-38°
      & 6$\times$ surround-view cameras; FOV 110°; 1280$\times$ 720 resolution; positions as in Figure~\ref{fig:anno} (c)
      & GNSS, IMU \\
    RSU & 
      1$\times$ 360° 64-channel LiDAR; 20 Hz rotation; Vertical FOV +10°/-38° 
      & 4$\times$ cameras (front, left, right, back); FOV 110°; 1280$\times$ 720 resolution 
      & GNSS \\
    Drone & 
      1$\times$ 360° 64-channel LiDAR; 20 Hz rotation; Vertical FOV -30°/-90° 
      & 1$\times$ downward-facing camera; FOV 110°; 1280$\times$ 720 resolution 
      & GNSS, IMU \\
    \bottomrule
  \end{tabular}
\end{table}

\subsection{Data Annotation and Downstream Tasks}
\label{sec:annotation}

\begin{figure}
  \centering
  \includegraphics[width=\linewidth]{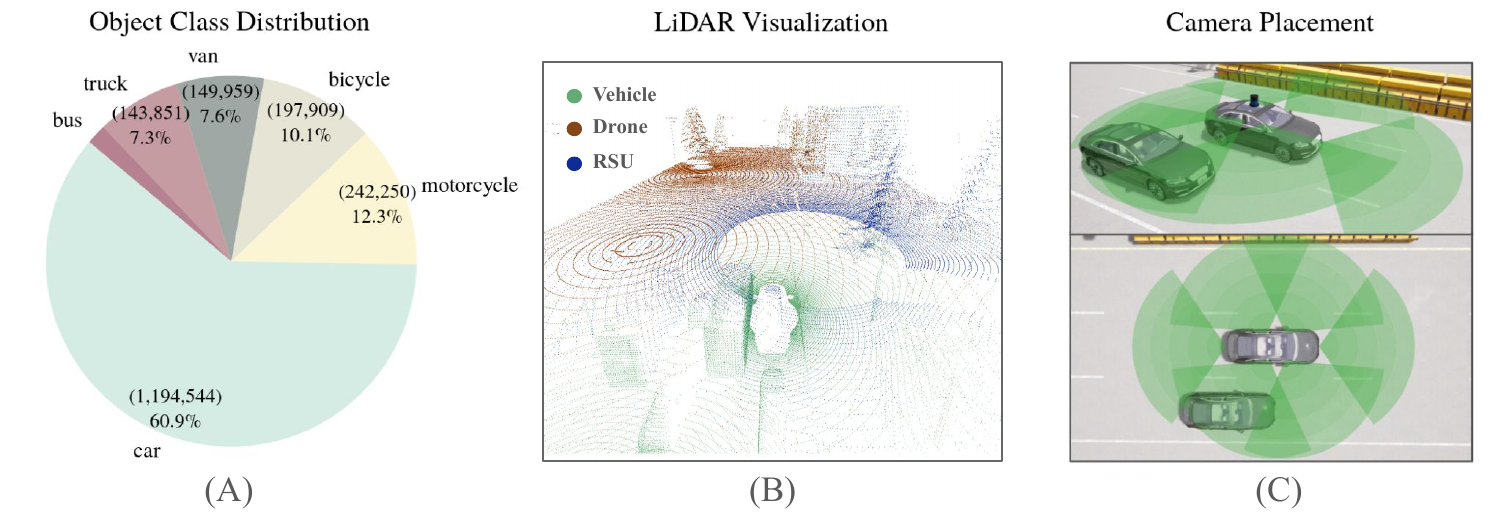}
  \caption{(a) The percentage of each scene in the dataset; (b) LiDAR point clouds visualization for each agent types, and (c) the camera/LiDAR placement for vehicular agents.}
  \label{fig:anno}
\end{figure}

CARLA simulator provides accurate ground truth labels for all the objects in the scene, including the 3D bounding boxes, semantic segmentation, depth map, and semantic LiDAR point clouds. Using such labels, we provide 3D bounding boxes annotations for 3D object detection, segmentation map annotations for BEV semantic segmentation, depth map annotations for depth estimation, and tracking annotations for 3D multi-object tracking. The data and annotations are gathered in a synchronized manner across all agents in the scene, enabling the collaborative perception tasks. The dataset is split into training, validation, and test sets that contain 2.19, 1.02, and 3.52 hours of driving sequences, respectively. 
We annotate 6 categories of common objects in the driving scenarios including cars, motorcycles, bicycles, vans, trucks, and buses, forming a total of 1,961,484 annotated objects. The statistics of the dataset are summarized in Figure \ref{fig:anno} (a).

\label{sec:dataset_statistics}
\section{Benchmark}

In this section, we benchmark multi-agent collaborative perception algorithms on our proposed AirV2X-Perception dataset. We carefully select six representative algorithms that showcase different approaches to collaborative perception challenges, with particular attention to their applicability in complex, heterogeneous, multi-agent scenarios involving ground vehicles, infrastructure, and aerial perspectives. Our evaluation spans 3D object detection, BEV semantic segmentation, and computational efficiency across diverse environmental conditions and agent configurations, providing insights into both algorithmic performance and practical deployment considerations.

\subsection{Experimental Setup and Methodology}

The AirV2X-Perception dataset features three agent types—vehicles, roadside units (RSUs), and drones—each with unique sensor configurations and placement characteristics. To adequately evaluate performance across this heterogeneous network, we selected algorithms representing various design philosophies in collaborative perception: Transformer-based approaches like \textbf{V2XViT}~\cite{xu2022v2x} that leverage transformer architectures proven effective in numerous computer vision tasks and demonstrate scalability to large datasets; communication-efficient frameworks such as \textbf{When2com}~\cite{liu2020when2com} and \textbf{Where2comm}~\cite{hu2022where2comm} that focus on optimizing communication efficiency—a critical consideration for large-scale connected agent deployments representative of real-world scenarios; BEV-Segmentation-optimized methods like \textbf{CoBEVT}~\cite{xu2022cobevt} that provide specialized capabilities for bird's-eye-view representations, applicable to both semantic segmentation and object detection tasks; and heterogeneous agent collaboration approaches including \textbf{HEAL}~\cite{lu2024extensible} and \textbf{STAMP}~\cite{gao2025stamp} that specifically address the challenges of collaboration between agents with differing sensor modalities, with HEAL employing backward alignment to synchronize heterogeneous agent models and STAMP optimizing efficiency through an adapter-reverter architecture.

All experiments were conducted under consistent hardware and software conditions using RTX A6000 GPUs with PyTorch 2.6.0, CUDA 12.6, and spconv 2.1. To ensure a fair comparison and maintain focus on core algorithm capabilities, this benchmark concentrates on LiDAR-based V2X perception, which is supported by most existing V2X perception algorithms and provides a common foundation for evaluation. We maintain consistent training hyperparameters across all methods where possible, including learning rate schedules, optimization algorithms, and data preprocessing techniques. For reproducibility and detailed implementation information, readers are directed to the accompanying codebase in the \href{https://github.com/taco-group/AirV2X-Perception}{AirV2X-Perception repository}\footnote{https://github.com/taco-group/AirV2X-Perception}.

\subsection{Performance Overview}

We evaluate each method along three primary dimensions: 3D object detection accuracy, BEV semantic segmentation quality, and computational efficiency (peak GPU memory usage). These dimensions collectively provide insight into both the perceptual capabilities and practical deployment considerations for these algorithms in real-world V2X scenarios. For 3D object detection, we establish a standardized evaluation area in the $x$ and $y$ directions spanning [-140.8, 140.8] meters and [-40, 40] meters with respect to the ego vehicle. We employ mean average precision (mAP) at 30\% and 50\% IoU thresholds (AP30 and AP50) as our primary metrics. For BEV semantic segmentation, we narrow the evaluation range to [-64, 64] meters and [-40, 40] meters in the $x$ and $y$ directions, respectively. We use mean intersection over union (mIoU) as the standard metric for segmentation performance. Following conventions established in prior datasets such as Dair-V2X~\cite{yu2022dair} and TUMTraf~\cite{zimmer2024tumtraf}, we categorize different vehicle types as a single class to maintain consistency with existing benchmarks. The peak GPU memory measurement indicates the maximum GPU memory consumption during training for each batch. Table \ref{tab:performance_overview} summarizes these evaluation results across all benchmark dimensions.

\begin{wraptable}{r}{0.5\textwidth}
  \centering
  \vspace{-4.5mm}
  \caption{Performance overview of 3D object detection, BEV semantic segmentation, and peak GPU memory usage.}
  \small
\resizebox{\linewidth}{!}{%
  \begin{tabular}{@{}l|l|cc|c@{}}
    \toprule
    Method 
    & {\scriptsize AP30$\uparrow$}  & {\scriptsize AP50$\uparrow$} & {\scriptsize mIoU$\uparrow$} & {\scriptsize VRAM$\downarrow${\tiny (GB)}}\\
    \midrule
    When2com ~\cite{liu2020when2com}      & 23.0 & 20.5 & 15.0 & \textbf{8.2} \\
    CoBEVT ~\cite{xu2022cobevt}             & 42.9 & 29.8 & 33.9 & 38.1\\
    Where2comm ~\cite{hu2022where2comm}  & 44.8 & 37.0 & 29.2 & 10.2 \\
    V2XViT ~\cite{xu2022v2x}             & 46.4 & 39.1 & 32.7 & 43.5 \\
    HEAL ~\cite{lu2024extensible}         & \textbf{49.2} & \textbf{45.5} & \textbf{33.9} & 12.4 \\
    STAMP ~\cite{gao2025stamp}          & \underline{47.9} & \underline{42.7} & 27.7 & \underline{10.1} \\
    \bottomrule
  \end{tabular}
}\label{tab:performance_overview}
\end{wraptable}

Our results reveal several important trends. First, \textbf{heterogeneous agent collaboration methods (HEAL and STAMP) demonstrate superior performance} in object detection tasks, with HEAL achieving the highest AP30 and AP50 scores of 49.2\% and 45.5\%, respectively. This suggests that explicit modeling of agent heterogeneity is beneficial in complex multi-agent scenarios like those in our AirV2X-Perception dataset. Second, we observe a clear \textbf{trade-off between performance and computational efficiency}. While HEAL achieves state-of-the-art detection accuracy, it requires 51\% more memory than When2com, the most memory-efficient approach. Interestingly, STAMP provides a compelling balance, with detection performance within 1.3-2.8 percentage points of HEAL while maintaining a favorable memory footprint (10.1 GB vs. 12.4 GB). For semantic segmentation, both HEAL and CoBEVT achieve the highest mIoU of 33.9\%, with V2XViT following closely at 32.7\%. The consistent strong performance of HEAL across both tasks highlights the importance of effectively handling heterogeneous agent inputs in collaborative perception systems.

\subsection{Performance Analysis Across Environmental Conditions}

To thoroughly evaluate real-world applicability, we analyze algorithm performance across various environmental factors: lighting conditions (day, dusk, or night), environments (urban or rural), and weather (rainy, foggy, cloudy, or clear). Table \ref{tab:environment_detection} presents these results for object detection tasks.

\begin{table}[htbp]
  \centering
  \scriptsize
  \setlength{\tabcolsep}{2.4pt}
  \caption{3D object detection results in different scenarios, including lighting conditions (day, dusk, or night), environments (urban or rural), and weather conditions (rainy, foggy, cloudy, or clear).}
  \begin{tabular}{l|cccccc|cccc|cccccccc}
    \toprule
    & \multicolumn{2}{c}{Day}
    & \multicolumn{2}{c}{Dusk}
    & \multicolumn{2}{c|}{Night}
    & \multicolumn{2}{c}{Urban}
    & \multicolumn{2}{c|}{Rural}
    & \multicolumn{2}{c}{Rainy}
    & \multicolumn{2}{c}{Foggy}
    & \multicolumn{2}{c}{Cloudy}
    & \multicolumn{2}{c}{Clear} \\
    \cmidrule(lr){2-3}  \cmidrule(lr){4-5}  \cmidrule(lr){6-7}
    \cmidrule(lr){8-9}  \cmidrule(lr){10-11} \cmidrule(lr){12-13}
    \cmidrule(lr){14-15} \cmidrule(lr){16-17} \cmidrule(lr){18-19}
    Method
    & {\tiny AP30} & {\tiny AP50} & {\tiny AP30} & {\tiny AP50} & {\tiny AP30} & {\tiny AP50} & {\tiny AP30} & {\tiny AP50}
    & {\tiny AP30} & {\tiny AP50} & {\tiny AP30} & {\tiny AP50} & {\tiny AP30} & {\tiny AP50} & {\tiny AP30} & {\tiny AP50}
    & {\tiny AP30} & {\tiny AP50} \\
    \midrule
    When2com~\cite{liu2020when2com}   & 21.5 & 20.8 & 24.8 & 20.6 & 15.4 & 15.0 & 25.3 & 22.4 & 16.1 & 15.5 & 17.2 & 16.4 & 32.7 & 26.4 & 25.3 & 22.4 & 22.7 & 22.1 \\
    CoBEVT~\cite{xu2022cobevt}      & 37.1 & 24.2 & 48.4 & 35.1 & 10.5 & 2.0 & 51.3 & 37.4 & 13.8 & 3.5 & 15.7 & 4.7 & 27.7 & 13.6 & 51.3 & 37.4 & 49.4 & 35.7 \\
    Where2comm~\cite{hu2022where2comm}   & 42.5 & 35.7 & 47.9 & 38.6 & 23.4 & 21.4 & 48.6 & 40.3 & 32.6 & 26.0 & 35.0 & 27.5 & 31.4 & 26.0 & 48.6 & 40.3 & 45.4 & 39.3 \\
    V2XViT~\cite{xu2022v2x}      & 46.8 & 41.1 & 47.8 & 39.0 & 19.3 & 16.2 & 52.1 & 43.8 & 28.7 & 25.1 & 32.3 & 28.6 & 38.4 & 31.3 & 52.1 & 43.8 & 50.7 & 44.3 \\
    HEAL~\cite{lu2024extensible}      & \textbf{49.2} & \textbf{46.9} & \textbf{49.9} & \textbf{44.7} & \textbf{32.8} & \textbf{31.7} & \textbf{52.8} & \textbf{48.9} & \textbf{36.8} & \textbf{34.1} & \textbf{38.5} & \textbf{35.7} & \textbf{45.6} & \textbf{39.3} & \textbf{52.8} & \textbf{48.9} & \textbf{54.8} & \textbf{52.8} \\
    STAMP~\cite{gao2025stamp}       & 48.6 & 41.6 & 48.0 & 39.7 & 19.8 & 16.9 & \textbf{52.8} & 44.5 & 28.6 & 25.7 & 32.0 & 29.2 & 38.5 & 31.6 & \textbf{52.8} & 44.5 & 52.1 & 45.0 \\
    \bottomrule
  \end{tabular}
  \label{tab:environment_detection}
\end{table}

We first observe that the performance of \textbf{all methods are greatly affected by the lighting conditions}-they show substantial performance degradation in nighttime scenarios compared to day and dusk conditions. CoBEVT experiences the most dramatic drop, with night AP30 (10.5\%) representing only 21.7\% of its dusk performance (48.4\%). HEAL demonstrates the highest resilience to lighting variations, maintaining 65.7\% of its daytime AP30 performance during nighttime conditions (32.8\% vs. 49.2\%). \textbf{Environmental context} also has great influence on the model performance. Urban environments consistently yield better detection results than rural settings across all methods. This performance gap is most pronounced for CoBEVT (51.3\% urban AP30 vs. 13.8\% rural AP30, a 73.1\% reduction) and least severe for HEAL (52.8\% vs. 36.8\%, a 30.3\% reduction). The heterogeneous information fusion approach of HEAL appears particularly valuable in rural environments where individual agent perspectives may be more limited. \textbf{Performance in adverse weather} presents interesting patterns. Foggy conditions seem to particularly challenge CoBEVT (27.7\% AP30) compared to clear conditions (49.4\% AP30), while HEAL maintains more consistent performance across weather variations. Rainy conditions degrade performance across all methods, though HEAL and STAMP show greater robustness than earlier approaches like When2com and CoBEVT. These findings highlight that collaborative perception algorithms differ in their ability to handle environmental variations, with heterogeneous-focused methods generally demonstrating greater robustness.

\subsection{Impact of drone navigation strategies}

A unique aspect of our AirV2X-Perception dataset is the incorporation of aerial agents (drones) with distinct navigation strategies: hover, patrol, and escort. We evaluate how these navigation strategies affect perception performance in Table \ref{tab:drone_detection}.

\begin{wraptable}{r}{0.5\textwidth}
  \centering
  \small
  \vspace{-4mm}
  \setlength{\tabcolsep}{2.4pt}
  \caption{Object detection results with different navigation strategies for drones, including hover, patrol, and escort.}
  \vspace{-2mm}
  \begin{tabular}{l|cc|cc|cc}
    \toprule
    & \multicolumn{2}{c|}{Hover}
    & \multicolumn{2}{c|}{Patrol}
    & \multicolumn{2}{c}{Escort} \\
    \cmidrule(lr){2-3} \cmidrule(lr){4-5} \cmidrule(lr){6-7}
    Method
    & {\scriptsize AP30} &  {\scriptsize AP50} & {\scriptsize AP30} &  {\scriptsize AP50} & {\scriptsize AP30} &  {\scriptsize AP50} \\
    \midrule
    When2com~\cite{liu2020when2com}  & 15.4 & 15.0 & 21.7 & 21.0 & 25.0 & 20.6 \\
    CoBEVT~\cite{xu2022cobevt}      & 10.5 & 2.0 & \textbf{64.2} & \textbf{54.3} & 21.5 & 9.0 \\
    Where2comm~\cite{hu2022where2comm}  & 23.4 & 21.4 & 57.8 & 48.4 & 31.9 & 25.7 \\
    V2XViT~\cite{xu2022v2x}      & 19.3 & 16.2 & 58.8 & 50.9 & 34.7 & 28.8 \\
    HEAL~\cite{lu2024extensible}        & \textbf{32.8} & \textbf{31.7} & 56.1 & 53.8 & \textbf{41.8} & \textbf{36.6} \\
    STAMP~\cite{gao2025stamp}       & 19.8 & 16.9 & 59.8 & 51.7 & 34.7 & 29.2 \\
    \bottomrule
  \end{tabular}
  \label{tab:drone_detection}
\end{wraptable}

According to Table \ref{tab:drone_detection}, we can observe that hover mode presents the greatest challenge for all methods, possibly due to the limited range covered by static drones. Patrol mode yields the best results across algorithms, particularly for CoBEVT (64.2\% AP30). Escort mode shows intermediate performance, with HEAL (41.8\% AP30) outperforming V2XViT and STAMP (both 34.7\% AP30), suggesting that more sophisticated fusion mechanisms better handle adaptive trajectories. These findings emphasize that drone navigation strategies should be coordinated with perception algorithm capabilities to optimize collaborative perception systems.

\subsection{Semantic Segmentation Performance Analysis}

The experiemental results of BEV semantic segmentation performance across environmental conditions and drone navigation strategies are shown in Table \ref{tab:environment_segmentation} and Table \ref{tab:drone_segmentation}, respectively.

\begin{wraptable}{r}{0.5\textwidth}
  \centering
  \small
  \setlength{\tabcolsep}{7.2pt}
  \vspace{-4mm}
  \caption{Semantic segmentation performance (mIoU) across various drone navigation strategies}
  \vspace{-2mm}
  \resizebox{\linewidth}{!}{
  \begin{tabular}{l|c|c|c}
    \toprule
    Method & Hover & Patrol & Escort \\
    \midrule
    When2com~\cite{liu2020when2com}   & 15.1 & 14.8 & 15.8 \\
    CoBEVT~\cite{xu2022cobevt}      & 34.3 & 40.3 & 28.4 \\
    Where2comm~\cite{hu2022where2comm}  & 31.7 & 33.8 & 27.6 \\
    V2XViT~\cite{xu2022v2x}      & 32.7 & \textbf{40.5} & \textbf{38.9} \\
    HEAL~\cite{lu2024extensible}        & \textbf{34.3} & 40.3 & 38.4 \\
    STAMP~\cite{gao2025stamp}       & 26.3 & 32.1 & 30.7 \\
    \bottomrule
  \end{tabular}
  }
  \vspace{-3.5mm}
  \label{tab:drone_segmentation}
\end{wraptable}

According to the results, segmentation performance demonstrates greater consistency across environmental conditions than object detection. V2XViT's segmentation performance varies by only 6.6\% points between its best (cloudy at 36.4\%) and worst (foggy at 28.8\%) conditions, compared to a 32.8\% point spread in detection between urban settings (52.1\%) and night scenarios (19.3\%). Different algorithms exhibit distinct environmental strengths. V2XViT achieves its highest segmentation mIoU of 34.7\% in daytime conditions. STAMP performs best in nighttime and cloudy scenarios with 35.4\% and 36.1\% mIoU respectively. CoBEVT excels in urban environments with 35.8\% mIoU, while HEAL demonstrates superior performance in foggy conditions at 30.7\% mIoU. For various drone navigation strategies, patrol mode yields the highest performance across all methods, with V2XViT reaching 40.5\% and both CoBEVT and HEAL achieving 40.3\%, consistent with detection results.

\begin{table}[h]
  \centering
  \small
  \setlength{\tabcolsep}{6.8pt}
  \caption{Segmentation performance (mIoU) across various lighting, environments, and weathers.}
  \resizebox{\columnwidth}{!}{%
  \begin{tabular}{@{}l|cc|cc|ccccc@{}}
    \toprule
    Method & Day & Dusk & Night & Urban & Rural & Rainy & Foggy & Cloudy & Clear \\
    \midrule
    When2com~\cite{liu2020when2com}   & 16.2 & 14.3 & 15.1 & 17.9 & 13.6 & 11.5 & 14.0 & 17.9 & 16.4 \\
    CoBEVT~\cite{xu2022cobevt}      & 33.1 & 34.2 & 34.3 & 33.7 & 27.0 & 28.3 & \textbf{30.7} & 33.7 & 32.7 \\
    Where2comm~\cite{hu2022where2comm}  & 32.8 & \textbf{36.5} & 31.7 & 31.6 & \textbf{28.6} & 24.1 & 28.7 & 31.6 & 30.0 \\
    V2XViT~\cite{xu2022v2x}      & \textbf{34.7} & 35.0 & \textbf{32.7} & \textbf{36.4} & 28.1 & \textbf{29.7} & 28.8 & \textbf{36.4} & \textbf{33.9} \\
    HEAL~\cite{lu2024extensible}        & 33.1 & 34.2 & 34.3 & 33.7 & 27.0 & 28.3 & \textbf{30.7} & 33.7 & 32.7 \\
    STAMP~\cite{gao2025stamp}       & 26.1 & 33.5 & 26.4 & 36.1 & 25.1 & 22.2 & 27.4 & 36.1 & 33.5 \\
    \bottomrule
  \end{tabular}
  }\label{tab:environment_segmentation}
\end{table}

\section{Discussion}

Our comprehensive evaluation of multi-agent collaborative perception algorithms reveals several key insights that can guide future research in both algorithmic improvement and dataset design.

\paragraph{Algorithmic Improvement.} 
The AirV2X-Perception dataset presents a particularly demanding benchmark due to its \textbf{heterogeneous collaboration requirements} among vehicles, roadside units (RSUs), and drones—each with distinct sensing capabilities and perspectives. Our results demonstrate that methods specifically designed for heterogeneous agents (HEAL and STAMP) consistently outperform conventional approaches, particularly in challenging environmental conditions. This underscores the importance of algorithms that can effectively integrate information from diverse sensing modalities and viewpoints. \textbf{Computational efficiency for large-scale collaborative perception} represents another significant challenge. Real-world deployments may involve dozens or hundreds of interconnected agents, yet several current approaches employ self-attention mechanisms that scale quadratically with agent count. The substantial variation in memory requirements across methods (8.2GB for When2com versus 43.5GB for V2XViT) highlights the need for algorithms that maintain perceptual accuracy while scaling efficiently to large agent networks. Furthermore, \textbf{performance robustness across diverse environmental conditions} remains problematic. Our analysis shows substantial performance degradation in challenging scenarios, with even the best-performing method (HEAL) experiencing a 34\% reduction in accuracy during nighttime operations. Developing environment-invariant collaborative perception systems that maintain consistent performance across all conditions constitutes a critical research direction.

\paragraph{Future Dataset Design.}
While the AirV2X-Perception dataset provides a comprehensive benchmark across various environments and agent configurations, \textbf{real-world deployment introduces additional complexities not fully captured in simulation}. Future research should prioritize datasets that incorporate realistic sensor noise, communication constraints, and environmental variations. Such datasets would enable more robust algorithm evaluation under authentic conditions and facilitate the development of methods that transfer effectively from simulation to real-world implementation. In practical applications, perception systems inform decision-making processes that subsequently affect future perceptions. \textbf{Developing closed-loop evaluation frameworks} that capture this feedback cycle would provide deeper insights into algorithm performance and enable optimization for long-term perception quality rather than merely single-frame accuracy. Finally, future datasets should \textbf{incorporate safety-critical edge cases} such as traffic accidents, road blockages, construction zones, and extreme weather events. These challenging scenarios often represent the most safety-critical situations for autonomous systems, making them particularly important for comprehensive evaluation and algorithm development.

\bibliography{reference}

\begin{thebibliography}{10}

\bibitem{zhang2023perception}
Yuxiao Zhang, Alexander Carballo, Hanting Yang, and Kazuya Takeda.
\newblock Perception and sensing for autonomous vehicles under adverse weather conditions: A survey.
\newblock {\em ISPRS Journal of Photogrammetry and Remote Sensing}, 196:146--177, 2023.

\bibitem{wang2025generative}
Yuping Wang, Shuo Xing, Cui Can, Renjie Li, Hongyuan Hua, Kexin Tian, Zhaobin Mo, Xiangbo Gao, Keshu Wu, Sulong Zhou, et~al.
\newblock Generative ai for autonomous driving: Frontiers and opportunities.
\newblock {\em arXiv preprint arXiv:2505.08854}, 2025.

\bibitem{gao2024mambast}
Xiangbo Gao, Asiegbu~Miracle Kanu-Asiegbu, and Xiaoxiao Du.
\newblock Mambast: A plug-and-play cross-spectral spatial-temporal fuser for efficient pedestrian detection.
\newblock {\em arXiv preprint arXiv:2408.01037}, 2024.

\bibitem{nokes2020v2i}
Tom Nokes, Ben Baxter, Harry Scammell, Denis Naberezhnykh, and Leonardo Provvedi.
\newblock Cost analysis of v2i deployment.
\newblock Technical Report ED13276, Issue No. 5, Ricardo Energy \& Environment, August 2020.

\bibitem{hu2022where2comm}
Yue Hu, Shaoheng Fang, Zixing Lei, Yiqi Zhong, and Siheng Chen.
\newblock Where2comm: Communication-efficient collaborative perception via spatial confidence maps.
\newblock {\em Advances in neural information processing systems}, 35:4874--4886, 2022.

\bibitem{gao2025stamp}
Xiangbo Gao, Runsheng Xu, Jiachen Li, Ziran Wang, Zhiwen Fan, and Zhengzhong Tu.
\newblock Stamp: Scalable task and model-agnostic collaborative perception.
\newblock {\em arXiv preprint arXiv:2501.18616}, 2025.

\bibitem{gao2025langcoop}
Xiangbo Gao, Yuheng Wu, Rujia Wang, Chenxi Liu, Yang Zhou, and Zhengzhong Tu.
\newblock Langcoop: Collaborative driving with language.
\newblock {\em arXiv preprint arXiv:2504.13406}, 2025.

\bibitem{wu2025synthetic}
Hanlin Wu, Pengfei Lin, Ehsan Javanmardi, Naren Bao, Bo~Qian, Hao Si, and Manabu Tsukada.
\newblock A synthetic benchmark for collaborative 3d semantic occupancy prediction in v2x autonomous driving.
\newblock {\em arXiv preprint arXiv:2506.17004}, 2025.

\bibitem{dosovitskiy2017carla}
Alexey Dosovitskiy, German Ros, Felipe Codevilla, Antonio Lopez, and Vladlen Koltun.
\newblock Carla: An open urban driving simulator.
\newblock In {\em Conference on robot learning}, pages 1--16. PMLR, 2017.

\bibitem{shah2018airsim}
Shital Shah, Debadeepta Dey, Chris Lovett, and Ashish Kapoor.
\newblock Airsim: High-fidelity visual and physical simulation for autonomous vehicles.
\newblock In {\em Field and Service Robotics: Results of the 11th International Conference}, pages 621--635. Springer, 2018.

\bibitem{xu2022opv2v}
Runsheng Xu, Hao Xiang, Xin Xia, Xu~Han, Jinlong Li, and Jiaqi Ma.
\newblock Opv2v: An open benchmark dataset and fusion pipeline for perception with vehicle-to-vehicle communication.
\newblock In {\em 2022 IEEE International Conference on Robotics and Automation (ICRA)}, pages 2583--2589, 2022.

\bibitem{li2022v2x}
Yiming Li, Dekun Ma, Ziyan An, Zixun Wang, Yiqi Zhong, Siheng Chen, and Chen Feng.
\newblock V2x-sim: Multi-agent collaborative perception dataset and benchmark for autonomous driving.
\newblock {\em IEEE Robotics and Automation Letters}, 7(4):10914--10921, 2022.

\bibitem{xu2022v2xvit}
Runsheng Xu, Hao Xiang, Zhengzhong Tu, Xin Xia, Ming-Hsuan Yang, and Jiaqi Ma.
\newblock V2x-vit: Vehicle-to-everything cooperative perception with vision transformer.
\newblock In {\em Computer Vision – ECCV 2022}, 2022.

\bibitem{yu2022dair}
Haibao Yu, Yizhen Luo, Mao Shu, Yiyi Huo, Zebang Yang, Yifeng Shi, Zhenglong Guo, Hanyu Li, Xing Hu, Jirui Yuan, and Zaiqing Nie.
\newblock Dair-v2x: A large-scale dataset for vehicle-infrastructure cooperative 3d object detection.
\newblock In {\em Proceedings of the IEEE/CVF Conference on Computer Vision and Pattern Recognition (CVPR)}, pages 21329--21338, 2022.

\bibitem{xu2023v2v4real}
Runsheng Xu, Xin Xia, Jinlong Li, Hanzhao Li, Shuo Zhang, Zhengzhong Tu, Zonglin Meng, Hao Xiang, Xiaoyu Dong, Rui Song, Hongkai Yu, Bolei Zhou, and Jiaqi Ma.
\newblock V2v4real: A real-world large-scale dataset for vehicle-to-vehicle cooperative perception.
\newblock In {\em Proceedings of the IEEE/CVF Conference on Computer Vision and Pattern Recognition (CVPR)}, pages 13712--13722, 2023.

\bibitem{hao2024rcooper}
Ruiyang Hao, Siqi Fan, Yingru Dai, Zhenlin Zhang, Chenxi Li, Yuntian Wang, Haibao Yu, Wenxian Yang, Jirui Yuan, and Zaiqing Nie.
\newblock Rcooper: A real-world large-scale dataset for roadside cooperative perception.
\newblock In {\em Proceedings of the IEEE/CVF Conference on Computer Vision and Pattern Recognition (CVPR)}, pages 22347--22357, 2024.

\bibitem{zimmer2024tumtraf}
Walter Zimmer, Gerhard~Arya Wardana, Suren Sritharan, Xingcheng Zhou, Rui Song, and Alois~C. Knoll.
\newblock Tumtraf v2x cooperative perception dataset.
\newblock In {\em Proceedings of the IEEE/CVF Conference on Computer Vision and Pattern Recognition (CVPR)}, pages 22668--22677, 2024.

\bibitem{ma2024holovic}
Cong Ma, Lei Qiao, Chengkai Zhu, Kai Liu, Zelong Kong, Qing Li, Xueqi Zhou, Yuheng Kan, and Wei Wu.
\newblock Holovic: Large-scale dataset and benchmark for multi-sensor holographic intersection and vehicle-infrastructure cooperative.
\newblock In {\em Proceedings of the IEEE/CVF Conference on Computer Vision and Pattern Recognition (CVPR)}, pages 22128--22137, 2024.

\bibitem{xiang2024v2xreal}
Hao Xiang, Zhaoliang Zheng, Xin Xia, Runsheng Xu, Letian Gao, Zewei Zhou, Xu~Han, Xinkai Ji, Mingxi Li, Zonglin Meng, Li~Jin, Mingyue Lei, Zhaoyang Ma, Zihang He, Haoxuan Ma, Yunshuang Yuan, Yingqian Zhao, and Jiaqi Ma.
\newblock V2x-real: A large-scale dataset for vehicle-to-everything cooperative perception.
\newblock In {\em Computer Vision – ECCV 2024}, pages 455--470, 2024.

\bibitem{huang2024v2xr}
Xun Huang, Jinlong Wang, Qiming Xia, Siheng Chen, Bisheng Yang, Xin Li, Cheng Wang, and Chenglu Wen.
\newblock V2x-r: Cooperative lidar-4d radar fusion for 3d object detection with denoising diffusion.
\newblock {\em arXiv preprint arXiv:2411.08402}, 2024.

\bibitem{hu2023collaboration}
Yue Hu, Yifan Lu, Runsheng Xu, Weidi Xie, Siheng Chen, and Yanfeng Wang.
\newblock Collaboration helps camera overtake lidar in 3d detection.
\newblock In {\em Proceedings of the IEEE/CVF Conference on Computer Vision and Pattern Recognition (CVPR)}, pages 9243--9252, 2023.

\bibitem{ye2024uav3d}
Hui Ye, Rajshekhar Sunderraman, and Shihao Ji.
\newblock Uav3d: A large-scale 3d perception benchmark for unmanned aerial vehicles.
\newblock In {\em Advances in Neural Information Processing Systems 37 (NeurIPS 2024) Datasets and Benchmarks Track}, 2024.

\bibitem{feng2024u2udata}
Tongtong Feng, Xin Wang, Feilin Han, Leping Zhang, and Wenwu Zhu.
\newblock U2udata: A large-scale cooperative perception dataset for swarm uavs autonomous flight.
\newblock In {\em Proceedings of the 32nd ACM International Conference on Multimedia (MM '24)}, 2024.

\bibitem{dutta2024mavrec}
Aritra Dutta, Srijan Das, Jacob Nielsen, Rajatsubhra Chakraborty, and Mubarak Shah.
\newblock Multiview aerial visual recognition (mavrec): Can multi-view improve aerial visual perception?
\newblock In {\em Proceedings of the IEEE/CVF Conference on Computer Vision and Pattern Recognition (CVPR)}, 2024.

\bibitem{wang2024uvcpnet}
Yuchao Wang, Peirui Cheng, Pengju Tian, Xiangru Li, Xiaoyu Zhang, and Licheng Jiao.
\newblock Uvcpnet: A uav-vehicle collaborative perception network for 3d object detection.
\newblock {\em arXiv preprint arXiv:2406.04647}, 2024.

\bibitem{wang2025griffin}
Jiahao Wang, Xiangyu Cao, Jiaru Zhong, Yuner Zhang, Haibao Yu, Lei He, and Shaobing Xu.
\newblock Griffin: Aerial-ground cooperative detection and tracking dataset and benchmark.
\newblock {\em arXiv preprint arXiv:2503.06983}, 2025.

\bibitem{hou2025agc}
Yunhao Hou, Bochao Zou, Min Zhang, Ran Chen, Shangdong Yang, Yanmei Zhang, Junbao Zhuo, Siheng Chen, Jiansheng Chen, and Huimin Ma.
\newblock Agc-drive: A large-scale dataset for real-world aerial-ground collaboration in driving scenarios.
\newblock {\em arXiv preprint arXiv:2506.16371}, 2025.

\bibitem{wang2025collaborative}
Naibang Wang, Deyong Shang, Yan Gong, Xiaoxi Hu, Ziying Song, Lei Yang, Yuhan Huang, Xiaoyu Wang, and Jianli Lu.
\newblock Collaborative perception datasets for autonomous driving: A review.
\newblock {\em arXiv preprint arXiv:2504.12696}, 2025.

\bibitem{gao2018object}
Hongbo Gao, Bo~Cheng, Jianqiang Wang, Keqiang Li, Jianhui Zhao, and Deyi Li.
\newblock Object classification using cnn-based fusion of vision and lidar in autonomous vehicle environment.
\newblock {\em IEEE Transactions on Industrial Informatics}, 14(9):4224--4231, 2018.

\bibitem{chen2019cooper}
Qi~Chen, Sihai Tang, Qing Yang, and Song Fu.
\newblock Cooper: Cooperative perception for connected autonomous vehicles based on 3d point clouds.
\newblock In {\em 2019 IEEE 39th International Conference on Distributed Computing Systems (ICDCS)}, pages 514--524. IEEE, 2019.

\bibitem{arnold2020cooperative}
Eduardo Arnold, Mehrdad Dianati, Robert de~Temple, and Saber Fallah.
\newblock Cooperative perception for 3d object detection in driving scenarios using infrastructure sensors.
\newblock {\em IEEE Transactions on Intelligent Transportation Systems}, 23(3):1852--1864, 2020.

\bibitem{melotti2020multimodal}
Gledson Melotti, Cristiano Premebida, and Nuno Gon{\c{c}}alves.
\newblock Multimodal deep-learning for object recognition combining camera and lidar data.
\newblock In {\em 2020 IEEE International Conference on Autonomous Robot Systems and Competitions (ICARSC)}, pages 177--182. IEEE, 2020.

\bibitem{fu2020depth}
Chen Fu, Chiyu Dong, Christoph Mertz, and John~M Dolan.
\newblock Depth completion via inductive fusion of planar lidar and monocular camera.
\newblock In {\em 2020 IEEE/RSJ International Conference on Intelligent Robots and Systems (IROS)}, pages 10843--10848. IEEE, 2020.

\bibitem{zeng2020dsdnet}
Wenyuan Zeng, Shenlong Wang, Renjie Liao, Yun Chen, Bin Yang, and Raquel Urtasun.
\newblock Dsdnet: Deep structured self-driving network.
\newblock In {\em Computer Vision--ECCV 2020: 16th European Conference, Glasgow, UK, August 23--28, 2020, Proceedings, Part XXI 16}, pages 156--172. Springer, 2020.

\bibitem{shi2022vips}
Shuyao Shi, Jiahe Cui, Zhehao Jiang, Zhenyu Yan, Guoliang Xing, Jianwei Niu, and Zhenchao Ouyang.
\newblock Vips: Real-time perception fusion for infrastructure-assisted autonomous driving.
\newblock In {\em Proceedings of the 28th annual international conference on mobile computing and networking}, pages 133--146, 2022.

\bibitem{glaser2023we}
Nathaniel~Moore Glaser and Zsolt Kira.
\newblock We need to talk: Identifying and overcoming communication-critical scenarios for self-driving.
\newblock {\em arXiv preprint arXiv:2305.04352}, 2023.

\bibitem{wang2020v2vnet}
Tsun-Hsuan Wang, Sivabalan Manivasagam, Ming Liang, Bin Yang, Wenyuan Zeng, and Raquel Urtasun.
\newblock V2vnet: Vehicle-to-vehicle communication for joint perception and prediction.
\newblock In {\em Computer Vision--ECCV 2020: 16th European Conference, Glasgow, UK, August 23--28, 2020, Proceedings, Part II 16}, pages 605--621. Springer, 2020.

\bibitem{liu2020when2com}
Yen-Cheng Liu, Junjiao Tian, Nathaniel Glaser, and Zsolt Kira.
\newblock When2com: Multi-agent perception via communication graph grouping.
\newblock In {\em Proceedings of the IEEE/CVF Conference on computer vision and pattern recognition}, pages 4106--4115, 2020.

\bibitem{cui2022coopernaut}
Jiaxun Cui, Hang Qiu, Dian Chen, Peter Stone, and Yuke Zhu.
\newblock Coopernaut: End-to-end driving with cooperative perception for networked vehicles.
\newblock In {\em Proceedings of the IEEE/CVF Conference on Computer Vision and Pattern Recognition}, pages 17252--17262, 2022.

\bibitem{xu2022v2x}
Runsheng Xu, Hao Xiang, Zhengzhong Tu, Xin Xia, Ming-Hsuan Yang, and Jiaqi Ma.
\newblock V2x-vit: Vehicle-to-everything cooperative perception with vision transformer.
\newblock In {\em European conference on computer vision}, pages 107--124. Springer, 2022.

\bibitem{qiao2023adaptive}
Donghao Qiao and Farhana Zulkernine.
\newblock Adaptive feature fusion for cooperative perception using lidar point clouds.
\newblock In {\em Proceedings of the IEEE/CVF winter conference on applications of computer vision}, pages 1186--1195, 2023.

\bibitem{li2023learning}
Jinlong Li, Runsheng Xu, Xinyu Liu, Jin Ma, Zicheng Chi, Jiaqi Ma, and Hongkai Yu.
\newblock Learning for vehicle-to-vehicle cooperative perception under lossy communication.
\newblock {\em IEEE Transactions on Intelligent Vehicles}, 8(4):2650--2660, 2023.

\bibitem{wang2023core}
Binglu Wang, Lei Zhang, Zhaozhong Wang, Yongqiang Zhao, and Tianfei Zhou.
\newblock Core: Cooperative reconstruction for multi-agent perception.
\newblock In {\em 2023 IEEE/CVF International Conference on Computer Vision (ICCV)}, pages 8676--8686. IEEE Computer Society, 2023.

\bibitem{yu2023vehicle}
Haibao Yu, Yingjuan Tang, Enze Xie, Jilei Mao, Jirui Yuan, Ping Luo, and Zaiqing Nie.
\newblock Vehicle-infrastructure cooperative 3d object detection via feature flow prediction.
\newblock {\em arXiv preprint arXiv:2303.10552}, 2023.

\bibitem{wang2025cocmt}
Rujia Wang, Xiangbo Gao, Hao Xiang, Runsheng Xu, and Zhengzhong Tu.
\newblock Cocmt: Communication-efficient cross-modal transformer for collaborative perception.
\newblock {\em arXiv preprint arXiv:2503.13504}, 2025.

\bibitem{luo2025senserag}
Xuewen Luo, Chenxi Liu, Fan Ding, Fengze Yang, Yang Zhou, Junnyong Loo, and Hwa~Hui Tew.
\newblock Senserag: Constructing environmental knowledge bases with proactive querying for llm-based autonomous driving.
\newblock In {\em Proceedings of the Winter Conference on Applications of Computer Vision}, pages 989--996, 2025.

\bibitem{you2024v2x}
Junwei You, Haotian Shi, Zhuoyu Jiang, Zilin Huang, Rui Gan, Keshu Wu, Xi~Cheng, Xiaopeng Li, and Bin Ran.
\newblock V2x-vlm: End-to-end v2x cooperative autonomous driving through large vision-language models.
\newblock {\em arXiv preprint arXiv:2408.09251}, 2024.

\bibitem{wu2025v2x}
Keshu Wu, Pei Li, Yang Zhou, Rui Gan, Junwei You, Yang Cheng, Jingwen Zhu, Steven~T Parker, Bin Ran, David~A Noyce, et~al.
\newblock V2x-llm: Enhancing v2x integration and understanding in connected vehicle corridors.
\newblock {\em arXiv preprint arXiv:2503.02239}, 2025.

\bibitem{luo2025v2x}
Xuewen Luo, Fengze Yang, Fan Ding, Xiangbo Gao, Shuo Xing, Yang Zhou, Zhengzhong Tu, and Chenxi Liu.
\newblock V2x-unipool: Unifying multimodal perception and knowledge reasoning for autonomous driving.
\newblock {\em arXiv preprint arXiv:2506.02580}, 2025.

\bibitem{xu2022cobevt}
Runsheng Xu, Zhengzhong Tu, Hao Xiang, Wei Shao, Bolei Zhou, and Jiaqi Ma.
\newblock Cobevt: Cooperative bird's eye view semantic segmentation with sparse transformers.
\newblock {\em arXiv preprint arXiv:2207.02202}, 2022.

\bibitem{chen2019f}
Qi~Chen, Xu~Ma, Sihai Tang, Jingda Guo, Qing Yang, and Song Fu.
\newblock F-cooper: Feature based cooperative perception for autonomous vehicle edge computing system using 3d point clouds.
\newblock In {\em Proceedings of the 4th ACM/IEEE Symposium on Edge Computing}, pages 88--100, 2019.

\bibitem{qu2024sicp}
Deyuan Qu, Qi~Chen, Tianyu Bai, Hongsheng Lu, Heng Fan, Hao Zhang, Song Fu, and Qing Yang.
\newblock Sicp: Simultaneous individual and cooperative perception for 3d object detection in connected and automated vehicles.
\newblock In {\em 2024 IEEE/RSJ International Conference on Intelligent Robots and Systems (IROS)}, pages 8905--8912. IEEE, 2024.

\bibitem{lu2024extensible}
Yifan Lu, Yue Hu, Yiqi Zhong, Dequan Wang, Yanfeng Wang, and Siheng Chen.
\newblock An extensible framework for open heterogeneous collaborative perception.
\newblock {\em arXiv preprint arXiv:2401.13964}, 2024.

\bibitem{GlobalDrone}
Philip~Butterworth Hayes.
\newblock {G}lobal drone industry market forecasts: analysts trim their growth predictions again - {U}nmanned airspace --- unmannedairspace.info.
\newblock [Accessed 17-05-2025].

\bibitem{afrin2024advancements}
Tanzina Afrin, Nita Yodo, Arup Dey, and Lucy~G Aragon.
\newblock Advancements in uav-enabled intelligent transportation systems: A three-layered framework and future directions.
\newblock {\em Applied Sciences}, 14(20):9455, 2024.

\bibitem{wang2025uav}
Dawei Wang and Ruonan Zhang.
\newblock Uav-assisted intelligent vehicular networks, 2025.

\bibitem{kavas2022v2x}
Ozgenur Kavas-Torris, Sukru~Yaren Gelbal, Mustafa~Ridvan Cantas, Bilin Aksun~Guvenc, and Levent Guvenc.
\newblock V2x communication between connected and automated vehicles (cavs) and unmanned aerial vehicles (uavs).
\newblock {\em Sensors}, 22(22):8941, 2022.

\bibitem{hadiwardoyo2019modelling}
Seilendria~Ardityarama Hadiwardoyo.
\newblock {\em Modelling and real deployment of c-its by integrating ground vehicles and unmanned aerial vehicles}.
\newblock PhD thesis, Universitat Polit{\`e}cnica de Val{\`e}ncia, 2019.

\bibitem{gupta2024latency}
Abhishek Gupta and Xavier~N Fernando.
\newblock Latency analysis of drone-assisted c-v2x communications for basic safety and co-operative perception messages.
\newblock {\em Drones (2504-446X)}, 8(10), 2024.

\end{thebibliography}
\clearpage

\appendix

\vspace{1ex} 

\renewcommand{\thefigure}{A\arabic{figure}}
\renewcommand{\thetable}{A\arabic{table}}
\setcounter{figure}{0}
\setcounter{table}{0}

\section{Experiments Details}

All experiments were conducted under consistent hardware and software conditions using RTX A6000 GPUs, PyTorch 2.6.0, CUDA 12.6, and SpConv 2.1. Models were trained with a batch size of 4 using the Adam optimizer with an initial learning rate of 0.002 for 20 epochs.
We trained and evaluated all models following their official implementations (or unofficial implementations where official ones were unavailable). Random seeds were fixed across all experiments to ensure reproducibility.
For additional implementation details, please refer to our open-source codebase\footnote{\url{https://github.com/taco-group/AirV2X-Perception}}.

\section{Ablation Studies For The Contribution of Different Types of Agents.}

\begin{table}[h]
  \centering
  \vspace{-4mm}
  \scriptsize
  \caption{Object Detection Results by Lighting Conditions Across Different Agent Combinations.}
  \setlength{\tabcolsep}{8.2pt}
  \resizebox{\linewidth}{!}{
  \begin{tabular}{l|cc|cc|cc|cc}
    \toprule
    & \multicolumn{2}{c|}{Overall}
    & \multicolumn{2}{c|}{Daytime}
    & \multicolumn{2}{c|}{Dusk}
    & \multicolumn{2}{c}{Nighttime} \\
    \cmidrule(lr){2-3} \cmidrule(lr){4-5} \cmidrule(lr){6-7} \cmidrule(lr){8-9}
    Method
    & {AP30} & {AP50} & {AP30} & {AP50} & {AP30} & {AP50} & {AP30} & {AP50} \\
    \midrule
    \multicolumn{9}{l}{\textbf{Vehicle + Infra + Drone}} \\
    \midrule
    When2comm   & 23.0 & 20.5 & 21.5 & 20.8 & 24.8 & 20.6 & 15.4 & 15.0 \\
    CoBEVT      & 42.9 & 29.8 & 37.1 & 24.2 & 48.4 & 35.1 & 10.5 & 2.0 \\
    Where2comm  & 44.8 & 37.0 & 42.5 & 35.7 & 47.9 & 38.6 & 23.4 & 21.4 \\
    V2XViT      & 46.4 & 39.1 & 46.8 & 41.1 & 47.8 & 39.0 & 19.3 & 16.2 \\
    HEAL        & 49.2 & 45.5 & 49.2 & 46.9 & 49.9 & 44.7 & 32.8 & 31.7 \\
    STAMP       & 47.9 & 42.7 & 47.6 & 41.6 & 48.0 & 39.7 & 19.8 & 16.9 \\
    \midrule
    \multicolumn{9}{l}{\textbf{Vehicle + Infra}} \\
    \midrule
    When2comm   & 16.2 {\textcolor{red}{\tiny 6.8$\downarrow$}} & 16.1 {\textcolor{red}{\tiny 4.4$\downarrow$}} & 16.4 {\textcolor{red}{\tiny 5.1$\downarrow$}} & 12.6 {\textcolor{red}{\tiny 8.2$\downarrow$}} & 19.0 {\textcolor{red}{\tiny 5.8$\downarrow$}} & 12.1 {\textcolor{red}{\tiny 8.5$\downarrow$}} & 7.9 {\textcolor{red}{\tiny 7.6$\downarrow$}} & 7.2 {\textcolor{red}{\tiny 7.9$\downarrow$}} \\
    CoBEVT      & 35.1 {\textcolor{red}{\tiny 7.8$\downarrow$}} & 21.7 {\textcolor{red}{\tiny 8.1$\downarrow$}} & 28.4 {\textcolor{red}{\tiny 8.7$\downarrow$}} & 19.4 {\textcolor{red}{\tiny 4.8$\downarrow$}} & 42.9 {\textcolor{red}{\tiny 5.5$\downarrow$}} & 31.2 {\textcolor{red}{\tiny 4.0$\downarrow$}} & 9.2 {\textcolor{red}{\tiny 1.3$\downarrow$}} & 0.7 {\textcolor{red}{\tiny 1.3$\downarrow$}} \\
    Where2comm  & 37.4 {\textcolor{red}{\tiny 7.4$\downarrow$}} & 32.9 {\textcolor{red}{\tiny 4.0$\downarrow$}} & 34.5 {\textcolor{red}{\tiny 8.0$\downarrow$}} & 31.6 {\textcolor{red}{\tiny 4.2$\downarrow$}} & 41.3 {\textcolor{red}{\tiny 6.6$\downarrow$}} & 35.0 {\textcolor{red}{\tiny 3.6$\downarrow$}} & 17.6 {\textcolor{red}{\tiny 5.8$\downarrow$}} & 14.8 {\textcolor{red}{\tiny 6.5$\downarrow$}} \\
    V2XViT      & 40.5 {\textcolor{red}{\tiny 5.9$\downarrow$}} & 32.8 {\textcolor{red}{\tiny 6.3$\downarrow$}} & 43.6 {\textcolor{red}{\tiny 3.2$\downarrow$}} & 35.1 {\textcolor{red}{\tiny 6.0$\downarrow$}} & 43.3 {\textcolor{red}{\tiny 4.5$\downarrow$}} & 35.4 {\textcolor{red}{\tiny 3.5$\downarrow$}} & 12.7 {\textcolor{red}{\tiny 6.6$\downarrow$}} & 12.3 {\textcolor{red}{\tiny 4.0$\downarrow$}} \\
    HEAL        & 45.3 {\textcolor{red}{\tiny 3.9$\downarrow$}} & 42.6 {\textcolor{red}{\tiny 3.0$\downarrow$}} & 44.5 {\textcolor{red}{\tiny 4.7$\downarrow$}} & 43.5 {\textcolor{red}{\tiny 3.4$\downarrow$}} & 46.4 {\textcolor{red}{\tiny 3.5$\downarrow$}} & 39.5 {\textcolor{red}{\tiny 5.2$\downarrow$}} & 22.8 {\textcolor{red}{\tiny 10.0$\downarrow$}} & 16.7 {\textcolor{red}{\tiny 15.0$\downarrow$}} \\
    STAMP       & 45.0 {\textcolor{red}{\tiny 2.9$\downarrow$}} & 40.8 {\textcolor{red}{\tiny 1.9$\downarrow$}} & 45.5 {\textcolor{red}{\tiny 2.0$\downarrow$}} & 40.0 {\textcolor{red}{\tiny 1.6$\downarrow$}} & 45.9 {\textcolor{red}{\tiny 2.1$\downarrow$}} & 38.0 {\textcolor{red}{\tiny 1.7$\downarrow$}} & 18.8 {\textcolor{red}{\tiny 1.0$\downarrow$}} & 15.0 {\textcolor{red}{\tiny 1.9$\downarrow$}} \\
    \midrule
    \multicolumn{9}{l}{\textbf{Vehicle only}} \\
    \midrule
    When2comm   & 11.3 {\textcolor{red}{\tiny 11.7$\downarrow$}} & 1.2 {\textcolor{red}{\tiny 19.3$\downarrow$}} & 10.6 {\textcolor{red}{\tiny 10.9$\downarrow$}} & 0.8 {\textcolor{red}{\tiny 20.0$\downarrow$}} & 12.0 {\textcolor{red}{\tiny 12.8$\downarrow$}} & 1.8 {\textcolor{red}{\tiny 18.9$\downarrow$}} & 3.0 {\textcolor{red}{\tiny 12.4$\downarrow$}} & 0.4 {\textcolor{red}{\tiny 14.7$\downarrow$}} \\
    CoBEVT      & 23.7 {\textcolor{red}{\tiny 19.2$\downarrow$}} & 8.0 {\textcolor{red}{\tiny 21.8$\downarrow$}} & 25.4 {\textcolor{red}{\tiny 11.7$\downarrow$}} & 5.1 {\textcolor{red}{\tiny 19.1$\downarrow$}} & 25.7 {\textcolor{red}{\tiny 22.7$\downarrow$}} & 5.5 {\textcolor{red}{\tiny 29.6$\downarrow$}} & 5.3 {\textcolor{red}{\tiny 5.2$\downarrow$}} & 0.5 {\textcolor{red}{\tiny 1.5$\downarrow$}} \\
    Where2comm  & 24.5 {\textcolor{red}{\tiny 20.3$\downarrow$}} & 5.3 {\textcolor{red}{\tiny 31.7$\downarrow$}} & 23.1 {\textcolor{red}{\tiny 19.4$\downarrow$}} & 4.7 {\textcolor{red}{\tiny 31.0$\downarrow$}} & 26.1 {\textcolor{red}{\tiny 21.8$\downarrow$}} & 5.8 {\textcolor{red}{\tiny 32.7$\downarrow$}} & 7.0 {\textcolor{red}{\tiny 16.4$\downarrow$}} & 2.4 {\textcolor{red}{\tiny 18.9$\downarrow$}} \\
    V2XViT      & 28.5 {\textcolor{red}{\tiny 17.9$\downarrow$}} & 9.3 {\textcolor{red}{\tiny 29.8$\downarrow$}} & 26.8 {\textcolor{red}{\tiny 20.0$\downarrow$}} & 9.6 {\textcolor{red}{\tiny 31.4$\downarrow$}} & 29.5 {\textcolor{red}{\tiny 18.2$\downarrow$}} & 9.2 {\textcolor{red}{\tiny 29.8$\downarrow$}} & 11.2 {\textcolor{red}{\tiny 8.0$\downarrow$}} & 1.8 {\textcolor{red}{\tiny 14.4$\downarrow$}} \\
    HEAL        & 32.4 {\textcolor{red}{\tiny 16.9$\downarrow$}} & 12.9 {\textcolor{red}{\tiny 32.7$\downarrow$}} & 29.7 {\textcolor{red}{\tiny 19.5$\downarrow$}} & 13.4 {\textcolor{red}{\tiny 33.5$\downarrow$}} & 33.3 {\textcolor{red}{\tiny 16.6$\downarrow$}} & 14.0 {\textcolor{red}{\tiny 30.7$\downarrow$}} & 12.5 {\textcolor{red}{\tiny 20.3$\downarrow$}} & 1.7 {\textcolor{red}{\tiny 30.0$\downarrow$}} \\
    STAMP       & 31.0 {\textcolor{red}{\tiny 16.9$\downarrow$}} & 13.4 {\textcolor{red}{\tiny 29.3$\downarrow$}} & 28.6 {\textcolor{red}{\tiny 18.9$\downarrow$}} & 13.8 {\textcolor{red}{\tiny 27.8$\downarrow$}} & 33.1 {\textcolor{red}{\tiny 14.9$\downarrow$}} & 12.7 {\textcolor{red}{\tiny 27.0$\downarrow$}} & 10.2 {\textcolor{red}{\tiny 9.6$\downarrow$}} & 1.8 {\textcolor{red}{\tiny 15.1$\downarrow$}} \\
    \bottomrule
  \end{tabular}
  }
  \label{tab:obj_lighting}
  \vspace{-2mm}
\end{table}

\begin{wraptable}{r}{0.5\textwidth}
  \centering
  \scriptsize
  \vspace{-4mm}
  \caption{Object Detection Results by Environments Across Different Agent Combinations.}
  \setlength{\tabcolsep}{4.8pt}
  \begin{tabular}{l|cc|cc}
    \toprule
    & \multicolumn{2}{c|}{Urban}
    & \multicolumn{2}{c}{Rural} \\
    \cmidrule(lr){2-3} \cmidrule(lr){4-5}
    Method
    & {AP30} & {AP50} & {AP30} & {AP50} \\
    \midrule
    \multicolumn{5}{l}{\textbf{Vehicle + Infra + Drone}} \\
    \midrule
    When2comm   & 25.3 & 22.4 & 16.1 & 15.5 \\
    CoBEVT      & 51.3 & 37.4 & 13.8 & 3.5 \\
    Where2comm  & 48.6 & 40.3 & 32.6 & 26.0 \\
    V2XViT      & 52.1 & 43.8 & 28.7 & 25.1 \\
    HEAL        & 52.8 & 48.9 & 36.8 & 34.1 \\
    STAMP       & 52.8 & 44.5 & 28.6 & 25.7 \\
    \midrule
    \multicolumn{5}{l}{\textbf{Vehicle + Infra}} \\
    \midrule
    When2comm   & 20.8 {\textcolor{red}{\tiny 4.4$\downarrow$}} & 15.0 {\textcolor{red}{\tiny 7.4$\downarrow$}} & 12.8 {\textcolor{red}{\tiny 3.3$\downarrow$}} & 10.0 {\textcolor{red}{\tiny 5.5$\downarrow$}} \\
    CoBEVT      & 47.2 {\textcolor{red}{\tiny 4.1$\downarrow$}} & 29.4 {\textcolor{red}{\tiny 8.1$\downarrow$}} & 8.9 {\textcolor{red}{\tiny 4.8$\downarrow$}} & 3.3 {\textcolor{red}{\tiny 0.1$\downarrow$}} \\
    Where2comm  & 40.1 {\textcolor{red}{\tiny 8.5$\downarrow$}} & 36.4 {\textcolor{red}{\tiny 3.9$\downarrow$}} & 26.5 {\textcolor{red}{\tiny 6.1$\downarrow$}} & 23.0 {\textcolor{red}{\tiny 3.0$\downarrow$}} \\
    V2XViT      & 44.6 {\textcolor{red}{\tiny 7.6$\downarrow$}} & 40.8 {\textcolor{red}{\tiny 3.1$\downarrow$}} & 22.9 {\textcolor{red}{\tiny 5.9$\downarrow$}} & 16.5 {\textcolor{red}{\tiny 8.6$\downarrow$}} \\
    HEAL        & 52.1 {\textcolor{red}{\tiny 0.8$\downarrow$}} & 46.5 {\textcolor{red}{\tiny 2.5$\downarrow$}} & 33.3 {\textcolor{red}{\tiny 3.4$\downarrow$}} & 30.5 {\textcolor{red}{\tiny 3.6$\downarrow$}} \\
    STAMP       & 52.3 {\textcolor{red}{\tiny 0.6$\downarrow$}} & 43.4 {\textcolor{red}{\tiny 1.0$\downarrow$}} & 25.9 {\textcolor{red}{\tiny 2.7$\downarrow$}} & 23.0 {\textcolor{red}{\tiny 2.7$\downarrow$}} \\
    \midrule
    \multicolumn{5}{l}{\textbf{Vehicle only}} \\
    \midrule
    When2comm   & 13.3 {\textcolor{red}{\tiny 12.0$\downarrow$}} & 1.4 {\textcolor{red}{\tiny 21.0$\downarrow$}} & 4.5 {\textcolor{red}{\tiny 11.6$\downarrow$}} & 0.4 {\textcolor{red}{\tiny 15.1$\downarrow$}} \\
    CoBEVT      & 32.0 {\textcolor{red}{\tiny 19.3$\downarrow$}} & 6.3 {\textcolor{red}{\tiny 31.1$\downarrow$}} & 6.7 {\textcolor{red}{\tiny 7.1$\downarrow$}} & 3.6 {\textcolor{blue}{\tiny 0.2$\uparrow$}} \\
    Where2comm  & 27.5 {\textcolor{red}{\tiny 21.1$\downarrow$}} & 6.1 {\textcolor{red}{\tiny 34.2$\downarrow$}} & 14.0 {\textcolor{red}{\tiny 18.6$\downarrow$}} & 2.7 {\textcolor{red}{\tiny 23.4$\downarrow$}} \\
    V2XViT      & 33.9 {\textcolor{red}{\tiny 18.3$\downarrow$}} & 10.8 {\textcolor{red}{\tiny 33.0$\downarrow$}} & 10.5 {\textcolor{red}{\tiny 18.3$\downarrow$}} & 4.7 {\textcolor{red}{\tiny 20.4$\downarrow$}} \\
    HEAL        & 37.8 {\textcolor{red}{\tiny 15.0$\downarrow$}} & 16.5 {\textcolor{red}{\tiny 32.4$\downarrow$}} & 12.5 {\textcolor{red}{\tiny 24.2$\downarrow$}} & 4.6 {\textcolor{red}{\tiny 29.5$\downarrow$}} \\
    STAMP       & 36.4 {\textcolor{red}{\tiny 16.5$\downarrow$}} & 16.2 {\textcolor{red}{\tiny 28.3$\downarrow$}} & 11.9 {\textcolor{red}{\tiny 16.8$\downarrow$}} & 3.3 {\textcolor{red}{\tiny 22.4$\downarrow$}} \\
    \bottomrule
  \end{tabular}
  \vspace{-4mm}
  \label{tab:obj_env}
\end{wraptable}

This section analyzes the impact of different agent types (vehicles, infrastructure, and drones) on perception performance across various environmental conditions and scenarios. 

As shown in Table \ref{tab:obj_lighting}, when examining overall performance, removing drone agents leads to a moderate performance drop, while using only vehicle agents results in a severe degradation. The performance decline is particularly pronounced in AP50 metrics, where vehicle-only configurations suffer decreases of up to 32.7\% for models like HEAL. This suggests that infrastructure agents provide critical spatial information that complements vehicle perspectives. For nighttime scenarios, the contribution of drone agents becomes especially valuable, with their removal causing HEAL's AP50 score to drop by 15.0\%. This highlights drones' ability to maintain visibility in low-light conditions where ground-based agents struggle. The aerial perspective provided by drones offers a strategic advantage in maintaining perception reliability across varying environmental conditions, particularly in scenarios with compromised lighting.

\begin{wraptable}{r}{0.6\textwidth}
  \centering
  \scriptsize
  \setlength{\tabcolsep}{1.8pt}
  \caption{Object Detection Results by Drones' Navigation Strategies Across Different Agent Combinations.}
  \begin{tabular}{l|cc|cc|cc}
    \toprule
    & \multicolumn{2}{c|}{Hover}
    & \multicolumn{2}{c|}{Patrol}
    & \multicolumn{2}{c}{Escort} \\
    \cmidrule(lr){2-3} \cmidrule(lr){4-5} \cmidrule(lr){6-7}
    Method
    & {\tiny AP30} & {\tiny AP50} & {\tiny AP30} & {\tiny AP50} & {\tiny AP30} & {\tiny AP50} \\
    \midrule
    \multicolumn{7}{l}{\textbf{Vehicle + Infra + Drone}} \\
    \midrule
    When2comm   & 15.4 & 15.0 & 21.7 & 21.0 & 25.0 & 20.6 \\
    CoBEVT      & 10.5 & 2.0 & 64.2 & 54.3 & 21.5 & 9.0 \\
    Where2comm  & 23.4 & 21.4 & 57.8 & 48.4 & 31.9 & 25.7 \\
    V2XViT      & 19.3 & 16.2 & 58.8 & 50.9 & 34.7 & 28.8 \\
    HEAL        & 32.8 & 31.7 & 56.1 & 53.8 & 41.8 & 36.6 \\
    STAMP       & 19.8 & 16.9 & 59.8 & 51.7 & 34.7 & 29.2 \\
    \midrule
    \multicolumn{7}{l}{\textbf{Vehicle + Infra}} \\
    \midrule
    When2comm   & 7.9 {\textcolor{red}{\tiny 7.6$\downarrow$}} & 7.2 {\textcolor{red}{\tiny 7.9$\downarrow$}} & 16.4 {\textcolor{red}{\tiny 5.3$\downarrow$}} & 13.8 {\textcolor{red}{\tiny 7.2$\downarrow$}} & 20.0 {\textcolor{red}{\tiny 5.0$\downarrow$}} & 15.6 {\textcolor{red}{\tiny 5.0$\downarrow$}} \\
    CoBEVT      & 8.2 {\textcolor{red}{\tiny 2.3$\downarrow$}} & 0.7 {\textcolor{red}{\tiny 1.3$\downarrow$}} & 57.1 {\textcolor{red}{\tiny 7.1$\downarrow$}} & 45.6 {\textcolor{red}{\tiny 8.7$\downarrow$}} & 15.1 {\textcolor{red}{\tiny 6.4$\downarrow$}} & 11.3 {\textcolor{blue}{\tiny 2.3$\uparrow$}} \\
    Where2comm  & 17.6 {\textcolor{red}{\tiny 5.8$\downarrow$}} & 14.8 {\textcolor{red}{\tiny 6.5$\downarrow$}} & 52.4 {\textcolor{red}{\tiny 5.4$\downarrow$}} & 42.3 {\textcolor{red}{\tiny 6.0$\downarrow$}} & 28.8 {\textcolor{red}{\tiny 3.0$\downarrow$}} & 21.3 {\textcolor{red}{\tiny 4.4$\downarrow$}} \\
    V2XViT      & 12.7 {\textcolor{red}{\tiny 6.6$\downarrow$}} & 12.3 {\textcolor{red}{\tiny 4.0$\downarrow$}} & 55.8 {\textcolor{red}{\tiny 3.0$\downarrow$}} & 44.9 {\textcolor{red}{\tiny 6.1$\downarrow$}} & 28.9 {\textcolor{red}{\tiny 5.8$\downarrow$}} & 20.4 {\textcolor{red}{\tiny 8.4$\downarrow$}} \\
    HEAL        & 22.8 {\textcolor{red}{\tiny 10.0$\downarrow$}} & 16.7 {\textcolor{red}{\tiny 15.0$\downarrow$}} & 57.6 {\textcolor{blue}{\tiny 1.5$\uparrow$}} & 49.0 {\textcolor{red}{\tiny 4.8$\downarrow$}} & 41.3 {\textcolor{red}{\tiny 0.5$\downarrow$}} & 36.4 {\textcolor{red}{\tiny 0.2$\downarrow$}} \\
    STAMP       & 18.8 {\textcolor{red}{\tiny 1.0$\downarrow$}} & 15.0 {\textcolor{red}{\tiny 1.9$\downarrow$}} & 58.6 {\textcolor{red}{\tiny 1.2$\downarrow$}} & 51.1 {\textcolor{red}{\tiny 0.6$\downarrow$}} & 35.1 {\textcolor{blue}{\tiny 0.5$\uparrow$}} & 27.2 {\textcolor{red}{\tiny 2.0$\downarrow$}} \\
    \midrule
    \multicolumn{7}{l}{\textbf{Vehicle only}} \\
    \midrule
    When2comm   & 3.0 {\textcolor{red}{\tiny 12.4$\downarrow$}} & 0.4 {\textcolor{red}{\tiny 14.7$\downarrow$}} & 16.1 {\textcolor{red}{\tiny 5.6$\downarrow$}} & 1.9 {\textcolor{red}{\tiny 19.2$\downarrow$}} & 6.5 {\textcolor{red}{\tiny 18.5$\downarrow$}} & 0.8 {\textcolor{red}{\tiny 19.8$\downarrow$}} \\
    CoBEVT      & 5.3 {\textcolor{red}{\tiny 5.2$\downarrow$}} & 0.5 {\textcolor{red}{\tiny 1.5$\downarrow$}} & 35.9 {\textcolor{red}{\tiny 28.2$\downarrow$}} & 13.5 {\textcolor{red}{\tiny 40.8$\downarrow$}} & 12.9 {\textcolor{red}{\tiny 8.7$\downarrow$}} & 3.5 {\textcolor{red}{\tiny 5.5$\downarrow$}} \\
    Where2comm  & 7.0 {\textcolor{red}{\tiny 16.4$\downarrow$}} & 2.4 {\textcolor{red}{\tiny 18.9$\downarrow$}} & 36.4 {\textcolor{red}{\tiny 21.4$\downarrow$}} & 9.8 {\textcolor{red}{\tiny 38.6$\downarrow$}} & 13.1 {\textcolor{red}{\tiny 18.8$\downarrow$}} & 1.6 {\textcolor{red}{\tiny 24.1$\downarrow$}} \\
    V2XViT      & 11.2 {\textcolor{red}{\tiny 8.0$\downarrow$}} & 1.8 {\textcolor{red}{\tiny 14.4$\downarrow$}} & 41.1 {\textcolor{red}{\tiny 17.8$\downarrow$}} & 15.3 {\textcolor{red}{\tiny 35.6$\downarrow$}} & 15.8 {\textcolor{red}{\tiny 18.9$\downarrow$}} & 4.7 {\textcolor{red}{\tiny 24.1$\downarrow$}} \\
    HEAL        & 12.5 {\textcolor{red}{\tiny 20.3$\downarrow$}} & 1.7 {\textcolor{red}{\tiny 30.0$\downarrow$}} & 45.5 {\textcolor{red}{\tiny 10.6$\downarrow$}} & 24.9 {\textcolor{red}{\tiny 28.9$\downarrow$}} & 18.0 {\textcolor{red}{\tiny 23.8$\downarrow$}} & 5.8 {\textcolor{red}{\tiny 30.8$\downarrow$}} \\
    STAMP       & 10.2 {\textcolor{red}{\tiny 9.6$\downarrow$}} & 1.8 {\textcolor{red}{\tiny 15.1$\downarrow$}} & 44.4 {\textcolor{red}{\tiny 15.4$\downarrow$}} & 22.7 {\textcolor{red}{\tiny 29.0$\downarrow$}} & 17.0 {\textcolor{red}{\tiny 17.7$\downarrow$}} & 4.6 {\textcolor{red}{\tiny 24.6$\downarrow$}} \\
    \bottomrule
  \end{tabular}
  \label{tab:obj_drone}
\end{wraptable}

Table \ref{tab:obj_env} shows that in urban settings, HEAL and STAMP demonstrate remarkable robustness to drone removal, with minimal AP30 drops of 0.8 and 0.6\% respectively, suggesting that the dense infrastructure in urban areas can partially compensate for the elevated perspective that drones provide. The multiple perception points available from infrastructure agents in urban environments appear sufficient to maintain reliable detection performance even without aerial data streams.
Rural environments tell a different story, with every agent type providing crucial information. The absence of dense infrastructure in rural settings makes drone perspectives particularly valuable, as evidenced by performance drops of 2.7\%-6.1\% AP30 when drones are removed. The vehicle-only configuration performs drastically worse in rural settings, with HEAL experiencing a 29.5\% decrease in AP50, highlighting the challenges vehicles face in rural perception without additional perspectives. This significant disparity underscores how the collaborative perception benefits vary substantially based on environmental context.

\begin{table}[h]
  \centering
  \caption{Object Detection Results by Weather Conditions Across Different Agent Combinations.}
\setlength{\tabcolsep}{8.2pt}
  \resizebox{\linewidth}{!}{
  \begin{tabular}{l|cc|cc|cc|cc}
    \toprule
    & \multicolumn{2}{c|}{Rainy}
    & \multicolumn{2}{c|}{Foggy}
    & \multicolumn{2}{c|}{Cloudy}
    & \multicolumn{2}{c}{Clear} \\
    \cmidrule(lr){2-3} \cmidrule(lr){4-5} \cmidrule(lr){6-7} \cmidrule(lr){8-9}
    Method
    & {AP30} & {AP50} & {AP30} & {AP50} & {AP30} & {AP50} & {AP30} & {AP50} \\
    \midrule
    \multicolumn{9}{l}{\textbf{Vehicle + Infra + Drone}} \\
    \midrule
    When2comm   & 17.2 & 16.4 & 32.7 & 26.4 & 25.3 & 22.4 & 22.7 & 22.1 \\
    CoBEVT      & 15.7 & 4.7 & 27.7 & 13.6 & 51.3 & 37.4 & 49.4 & 35.7 \\
    Where2comm  & 35.0 & 27.5 & 31.4 & 26.0 & 48.6 & 40.3 & 45.4 & 39.3 \\
    V2XViT      & 32.3 & 28.6 & 38.4 & 31.3 & 52.1 & 43.8 & 50.7 & 44.3 \\
    HEAL        & 38.5 & 35.7 & 45.6 & 39.3 & 52.8 & 48.9 & 54.8 & 52.8 \\
    STAMP       & 32.0 & 29.2 & 38.5 & 31.6 & 52.8 & 44.5 & 52.1 & 45.0 \\
    \midrule
    \multicolumn{9}{l}{\textbf{Vehicle + Infra}} \\
    \midrule
    When2comm   & 11.0 {\textcolor{red}{\tiny 6.2$\downarrow$}} & 9.5 {\textcolor{red}{\tiny 6.8$\downarrow$}} & 28.9 {\textcolor{red}{\tiny 3.8$\downarrow$}} & 20.0 {\textcolor{red}{\tiny 6.4$\downarrow$}} & 20.8 {\textcolor{red}{\tiny 4.4$\downarrow$}} & 15.0 {\textcolor{red}{\tiny 7.4$\downarrow$}} & 18.5 {\textcolor{red}{\tiny 4.3$\downarrow$}} & 16.9 {\textcolor{red}{\tiny 5.1$\downarrow$}} \\
    CoBEVT      & 13.5 {\textcolor{red}{\tiny 2.2$\downarrow$}} & 3.3 {\textcolor{red}{\tiny 1.4$\downarrow$}} & 23.2 {\textcolor{red}{\tiny 4.5$\downarrow$}} & 7.2 {\textcolor{red}{\tiny 6.4$\downarrow$}} & 47.2 {\textcolor{red}{\tiny 4.1$\downarrow$}} & 29.4 {\textcolor{red}{\tiny 8.1$\downarrow$}} & 42.6 {\textcolor{red}{\tiny 6.8$\downarrow$}} & 31.2 {\textcolor{red}{\tiny 4.5$\downarrow$}} \\
    Where2comm  & 27.6 {\textcolor{red}{\tiny 7.4$\downarrow$}} & 24.0 {\textcolor{red}{\tiny 3.5$\downarrow$}} & 22.8 {\textcolor{red}{\tiny 8.6$\downarrow$}} & 19.8 {\textcolor{red}{\tiny 6.2$\downarrow$}} & 40.1 {\textcolor{red}{\tiny 8.5$\downarrow$}} & 36.4 {\textcolor{red}{\tiny 3.9$\downarrow$}} & 39.7 {\textcolor{red}{\tiny 5.8$\downarrow$}} & 36.0 {\textcolor{red}{\tiny 3.4$\downarrow$}} \\
    V2XViT      & 26.6 {\textcolor{red}{\tiny 5.7$\downarrow$}} & 23.0 {\textcolor{red}{\tiny 5.6$\downarrow$}} & 35.2 {\textcolor{red}{\tiny 3.2$\downarrow$}} & 26.0 {\textcolor{red}{\tiny 5.4$\downarrow$}} & 44.6 {\textcolor{red}{\tiny 7.6$\downarrow$}} & 40.8 {\textcolor{red}{\tiny 3.1$\downarrow$}} & 46.4 {\textcolor{red}{\tiny 4.3$\downarrow$}} & 36.1 {\textcolor{red}{\tiny 8.2$\downarrow$}} \\
    HEAL        & 36.4 {\textcolor{red}{\tiny 2.1$\downarrow$}} & 33.8 {\textcolor{red}{\tiny 1.9$\downarrow$}} & 43.8 {\textcolor{red}{\tiny 1.8$\downarrow$}} & 38.6 {\textcolor{red}{\tiny 0.6$\downarrow$}} & 50.1 {\textcolor{red}{\tiny 2.8$\downarrow$}} & 46.5 {\textcolor{red}{\tiny 2.5$\downarrow$}} & 50.6 {\textcolor{red}{\tiny 4.3$\downarrow$}} & 46.6 {\textcolor{red}{\tiny 6.2$\downarrow$}} \\
    STAMP       & 30.2 {\textcolor{red}{\tiny 1.8$\downarrow$}} & 26.9 {\textcolor{red}{\tiny 2.4$\downarrow$}} & 36.7 {\textcolor{red}{\tiny 1.8$\downarrow$}} & 31.1 {\textcolor{red}{\tiny 0.5$\downarrow$}} & 52.3 {\textcolor{red}{\tiny 0.6$\downarrow$}} & 43.4 {\textcolor{red}{\tiny 1.0$\downarrow$}} & 50.5 {\textcolor{red}{\tiny 1.5$\downarrow$}} & 44.1 {\textcolor{red}{\tiny 0.9$\downarrow$}} \\
    \midrule
    \multicolumn{9}{l}{\textbf{Vehicle only}} \\
    \midrule
    When2comm   & 5.1 {\textcolor{red}{\tiny 12.1$\downarrow$}} & 0.4 {\textcolor{red}{\tiny 16.0$\downarrow$}} & 8.3 {\textcolor{red}{\tiny 24.4$\downarrow$}} & 1.2 {\textcolor{red}{\tiny 25.2$\downarrow$}} & 13.3 {\textcolor{red}{\tiny 12.0$\downarrow$}} & 1.4 {\textcolor{red}{\tiny 21.0$\downarrow$}} & 14.3 {\textcolor{red}{\tiny 8.4$\downarrow$}} & 1.1 {\textcolor{red}{\tiny 21.0$\downarrow$}} \\
    CoBEVT      & 5.5 {\textcolor{red}{\tiny 10.1$\downarrow$}} & 2.7 {\textcolor{red}{\tiny 2.0$\downarrow$}} & 15.2 {\textcolor{red}{\tiny 12.5$\downarrow$}} & 2.8 {\textcolor{red}{\tiny 10.8$\downarrow$}} & 32.0 {\textcolor{red}{\tiny 19.3$\downarrow$}} & 6.3 {\textcolor{red}{\tiny 31.1$\downarrow$}} & 32.0 {\textcolor{red}{\tiny 17.4$\downarrow$}} & 10.2 {\textcolor{red}{\tiny 25.5$\downarrow$}} \\
    Where2comm  & 15.9 {\textcolor{red}{\tiny 19.1$\downarrow$}} & 2.8 {\textcolor{red}{\tiny 24.7$\downarrow$}} & 13.0 {\textcolor{red}{\tiny 18.4$\downarrow$}} & 1.5 {\textcolor{red}{\tiny 24.5$\downarrow$}} & 27.5 {\textcolor{red}{\tiny 21.1$\downarrow$}} & 6.1 {\textcolor{red}{\tiny 34.2$\downarrow$}} & 25.7 {\textcolor{red}{\tiny 19.7$\downarrow$}} & 5.8 {\textcolor{red}{\tiny 33.6$\downarrow$}} \\
    V2XViT      & 10.8 {\textcolor{red}{\tiny 21.5$\downarrow$}} & 5.8 {\textcolor{red}{\tiny 22.8$\downarrow$}} & 21.1 {\textcolor{red}{\tiny 17.3$\downarrow$}} & 5.6 {\textcolor{red}{\tiny 25.8$\downarrow$}} & 33.9 {\textcolor{red}{\tiny 18.3$\downarrow$}} & 10.8 {\textcolor{red}{\tiny 33.0$\downarrow$}} & 36.6 {\textcolor{red}{\tiny 14.2$\downarrow$}} & 12.0 {\textcolor{red}{\tiny 32.3$\downarrow$}} \\
    HEAL        & 12.2 {\textcolor{red}{\tiny 26.3$\downarrow$}} & 5.0 {\textcolor{red}{\tiny 30.7$\downarrow$}} & 21.0 {\textcolor{red}{\tiny 24.6$\downarrow$}} & 8.0 {\textcolor{red}{\tiny 31.3$\downarrow$}} & 37.8 {\textcolor{red}{\tiny 15.0$\downarrow$}} & 16.5 {\textcolor{red}{\tiny 32.4$\downarrow$}} & 38.3 {\textcolor{red}{\tiny 16.6$\downarrow$}} & 19.1 {\textcolor{red}{\tiny 33.6$\downarrow$}} \\
    STAMP       & 12.8 {\textcolor{red}{\tiny 19.3$\downarrow$}} & 3.8 {\textcolor{red}{\tiny 25.4$\downarrow$}} & 21.1 {\textcolor{red}{\tiny 17.4$\downarrow$}} & 5.8 {\textcolor{red}{\tiny 25.7$\downarrow$}} & 36.4 {\textcolor{red}{\tiny 16.5$\downarrow$}} & 16.2 {\textcolor{red}{\tiny 28.3$\downarrow$}} & 37.4 {\textcolor{red}{\tiny 14.7$\downarrow$}} & 19.0 {\textcolor{red}{\tiny 26.0$\downarrow$}} \\
    \bottomrule
  \end{tabular}
  }
  \label{tab:obj_weather}
\end{table}

Table \ref{tab:obj_drone} displays the impact of different drone flight patterns on performance. In hover scenarios, drones provide essential overhead perspectives, with their removal causing significant decreases in AP50 (1.9\%-15.0\%). Vehicle-only configurations suffer catastrophic degradation in hover scenarios, with AP50 decreases of up to 30.0\% for HEAL. For patrol scenarios, where drones follow predetermined routes, HEAL surprisingly shows a 1.5-point AP30 improvement when removing drones, indicating potential conflicts between drone and ground agent information during this pattern. However, this anomaly is not reflected in AP50 metrics, where all models show performance drops. The escort pattern reveals interesting dynamics, with STAMP showing a slight 0.5\% improvement in AP30 when removing drones, suggesting that closely following drone patterns may sometimes introduce redundant or conflicting information.

\begin{table}[h]
  \centering
  \scriptsize
  \setlength{\tabcolsep}{2.2pt}
  \caption{Semantic Segmentation Results by Lighting Conditions, Environments, and Weather Conditions Across Different Agent Combinations.}
  \resizebox{\linewidth}{!}{
  \begin{tabular}{l|c|ccc|cc|cccc}
    \toprule
    Method & Overall & Day & Dusk & Night & Urban & Rural & Rainy & Foggy & Cloudy & Clear \\
    \midrule
    \multicolumn{11}{l}{\textbf{Vehicle + Infra + Drone}} \\
    \midrule
    When2comm   & 15.0 & 16.2 & 14.3 & 15.1 & 17.9 & 13.6 & 11.5 & 14.0 & 17.9 & 16.4 \\
    CoBEVT      & 33.9 & 33.1 & 34.2 & 34.3 & 33.7 & 27.0 & 28.3 & 30.7 & 33.7 & 32.7 \\
    Where2comm  & 29.2 & 32.8 & 36.5 & 31.7 & 31.6 & 28.6 & 24.1 & 28.7 & 31.6 & 30.0 \\
    V2XViT      & 32.7 & 34.7 & 35.0 & 32.7 & 36.4 & 28.1 & 29.7 & 28.8 & 36.4 & 33.9 \\
    HEAL        & 33.9 & 33.1 & 34.2 & 34.3 & 33.7 & 27.0 & 28.3 & 30.7 & 33.7 & 32.7 \\
    STAMP       & 27.7 & 26.1 & 33.5 & 26.3 & 36.1 & 25.1 & 22.2 & 27.4 & 36.1 & 33.5 \\
    \midrule
    \multicolumn{11}{l}{\textbf{Vehicle + Infra}} \\
    \midrule
    When2comm   & 13.6 {\textcolor{red}{\tiny 1.3$\downarrow$}} & 14.7 {\textcolor{red}{\tiny 1.4$\downarrow$}} & 13.1 {\textcolor{red}{\tiny 1.3$\downarrow$}} & 13.8 {\textcolor{red}{\tiny 1.3$\downarrow$}}  & 16.3 {\textcolor{red}{\tiny 1.6$\downarrow$}} & 12.3 {\textcolor{red}{\tiny 1.3$\downarrow$}} & 10.4 {\textcolor{red}{\tiny 1.0$\downarrow$}} & 12.7 {\textcolor{red}{\tiny 1.3$\downarrow$}} & 16.3 {\textcolor{red}{\tiny 1.6$\downarrow$}} & 14.9 {\textcolor{red}{\tiny 1.5$\downarrow$}} \\
    CoBEVT      & 27.9 {\textcolor{red}{\tiny 6.1$\downarrow$}} & 27.1 {\textcolor{red}{\tiny 6.0$\downarrow$}} & 28.0 {\textcolor{red}{\tiny 6.1$\downarrow$}} & 28.2 {\textcolor{red}{\tiny 6.2$\downarrow$}}  & 27.6 {\textcolor{red}{\tiny 6.1$\downarrow$}} & 22.1 {\textcolor{red}{\tiny 4.8$\downarrow$}} & 23.3 {\textcolor{red}{\tiny 5.1$\downarrow$}} & 25.2 {\textcolor{red}{\tiny 5.6$\downarrow$}} & 27.6 {\textcolor{red}{\tiny 6.1$\downarrow$}} & 26.8 {\textcolor{red}{\tiny 5.9$\downarrow$}} \\
    Where2comm  & 24.0 {\textcolor{red}{\tiny 5.2$\downarrow$}} & 26.9 {\textcolor{red}{\tiny 5.9$\downarrow$}} & 29.9 {\textcolor{red}{\tiny 6.6$\downarrow$}} & 26.0 {\textcolor{red}{\tiny 5.7$\downarrow$}}  & 25.9 {\textcolor{red}{\tiny 5.7$\downarrow$}} & 23.5 {\textcolor{red}{\tiny 5.2$\downarrow$}} & 19.8 {\textcolor{red}{\tiny 4.3$\downarrow$}} & 23.6 {\textcolor{red}{\tiny 5.2$\downarrow$}} & 25.9 {\textcolor{red}{\tiny 5.7$\downarrow$}} & 24.6 {\textcolor{red}{\tiny 5.4$\downarrow$}} \\
    V2XViT      & 26.8 {\textcolor{red}{\tiny 5.9$\downarrow$}} & 28.5 {\textcolor{red}{\tiny 6.3$\downarrow$}} & 28.7 {\textcolor{red}{\tiny 6.3$\downarrow$}} & 26.8 {\textcolor{red}{\tiny 5.9$\downarrow$}}  & 29.8 {\textcolor{red}{\tiny 6.5$\downarrow$}} & 23.1 {\textcolor{red}{\tiny 5.0$\downarrow$}} & 24.4 {\textcolor{red}{\tiny 5.3$\downarrow$}} & 23.6 {\textcolor{red}{\tiny 5.2$\downarrow$}} & 29.8 {\textcolor{red}{\tiny 6.5$\downarrow$}} & 27.8 {\textcolor{red}{\tiny 6.1$\downarrow$}} \\
    HEAL        & 27.9 {\textcolor{red}{\tiny 6.1$\downarrow$}} & 27.1 {\textcolor{red}{\tiny 6.0$\downarrow$}} & 28.1 {\textcolor{red}{\tiny 6.1$\downarrow$}} & 28.1 {\textcolor{red}{\tiny 6.2$\downarrow$}}  & 27.7 {\textcolor{red}{\tiny 6.1$\downarrow$}} & 22.1 {\textcolor{red}{\tiny 4.8$\downarrow$}} & 23.3 {\textcolor{red}{\tiny 5.1$\downarrow$}} & 25.2 {\textcolor{red}{\tiny 5.5$\downarrow$}} & 27.7 {\textcolor{red}{\tiny 6.1$\downarrow$}} & 26.8 {\textcolor{red}{\tiny 5.9$\downarrow$}} \\
    STAMP       & 22.7 {\textcolor{red}{\tiny 5.0$\downarrow$}} & 21.4 {\textcolor{red}{\tiny 4.7$\downarrow$}} & 27.5 {\textcolor{red}{\tiny 6.0$\downarrow$}} & 21.5 {\textcolor{red}{\tiny 4.7$\downarrow$}}  & 29.6 {\textcolor{red}{\tiny 6.5$\downarrow$}} & 20.6 {\textcolor{red}{\tiny 4.5$\downarrow$}} & 18.3 {\textcolor{red}{\tiny 4.0$\downarrow$}} & 22.5 {\textcolor{red}{\tiny 4.9$\downarrow$}} & 29.6 {\textcolor{red}{\tiny 6.5$\downarrow$}} & 27.5 {\textcolor{red}{\tiny 6.0$\downarrow$}} \\
    \midrule
    \multicolumn{11}{l}{\textbf{Vehicle only}} \\
    \midrule
    When2comm   & 12.5 {\textcolor{red}{\tiny 2.4$\downarrow$}} & 13.6 {\textcolor{red}{\tiny 2.6$\downarrow$}} & 12.1 {\textcolor{red}{\tiny 2.3$\downarrow$}}   & 12.7 {\textcolor{red}{\tiny 2.4$\downarrow$}}    & 15.0 {\textcolor{red}{\tiny 2.9$\downarrow$}} & 11.4 {\textcolor{red}{\tiny 2.2$\downarrow$}} & 9.6 {\textcolor{red}{\tiny 1.8$\downarrow$}} & 11.8 {\textcolor{red}{\tiny 2.2$\downarrow$}}      & 15.0 {\textcolor{red}{\tiny 2.9$\downarrow$}} & 13.8 {\textcolor{red}{\tiny 2.6$\downarrow$}} \\
    CoBEVT      & 23.7 {\textcolor{red}{\tiny 10.2$\downarrow$}} & 23.2 {\textcolor{red}{\tiny 9.9$\downarrow$}} & 23.9 {\textcolor{red}{\tiny 10.3$\downarrow$}} & 24.0 {\textcolor{red}{\tiny 10.3$\downarrow$}} & 23.6 {\textcolor{red}{\tiny 10.1$\downarrow$}} & 18.9 {\textcolor{red}{\tiny 8.1$\downarrow$}} & 19.8 {\textcolor{red}{\tiny 8.5$\downarrow$}} & 21.5 {\textcolor{red}{\tiny 9.2$\downarrow$}}  & 23.6 {\textcolor{red}{\tiny 10.1$\downarrow$}} & 22.9 {\textcolor{red}{\tiny 9.8$\downarrow$}} \\
    Where2comm  & 20.5 {\textcolor{red}{\tiny 8.8$\downarrow$}} & 22.9 {\textcolor{red}{\tiny 9.8$\downarrow$}} & 25.6 {\textcolor{red}{\tiny 10.9$\downarrow$}}  & 22.2 {\textcolor{red}{\tiny 9.5$\downarrow$}}   & 22.1 {\textcolor{red}{\tiny 9.5$\downarrow$}} & 20.0 {\textcolor{red}{\tiny 8.6$\downarrow$}} & 16.9 {\textcolor{red}{\tiny 7.3$\downarrow$}} & 20.1 {\textcolor{red}{\tiny 8.6$\downarrow$}}    & 22.1 {\textcolor{red}{\tiny 9.5$\downarrow$}} & 21.0 {\textcolor{red}{\tiny 9.0$\downarrow$}} \\
    V2XViT      & 22.9 {\textcolor{red}{\tiny 9.8$\downarrow$}} & 24.3 {\textcolor{red}{\tiny 10.4$\downarrow$}} & 24.4 {\textcolor{red}{\tiny 10.5$\downarrow$}} & 22.9 {\textcolor{red}{\tiny 9.8$\downarrow$}}  & 25.4 {\textcolor{red}{\tiny 10.9$\downarrow$}} & 19.7 {\textcolor{red}{\tiny 8.4$\downarrow$}} & 20.7 {\textcolor{red}{\tiny 8.9$\downarrow$}} & 20.2 {\textcolor{red}{\tiny 8.6$\downarrow$}} & 25.4 {\textcolor{red}{\tiny 10.9$\downarrow$}} & 23.7 {\textcolor{red}{\tiny 10.2$\downarrow$}} \\
    HEAL        & 23.8 {\textcolor{red}{\tiny 10.2$\downarrow$}} & 23.2 {\textcolor{red}{\tiny 9.9$\downarrow$}} & 23.9 {\textcolor{red}{\tiny 10.2$\downarrow$}} & 24.1 {\textcolor{red}{\tiny 10.3$\downarrow$}} & 23.6 {\textcolor{red}{\tiny 10.1$\downarrow$}} & 18.8 {\textcolor{red}{\tiny 8.1$\downarrow$}} & 19.8 {\textcolor{red}{\tiny 8.5$\downarrow$}} & 21.5 {\textcolor{red}{\tiny 9.2$\downarrow$}}  & 23.6 {\textcolor{red}{\tiny 10.1$\downarrow$}} & 22.9 {\textcolor{red}{\tiny 9.8$\downarrow$}} \\
    STAMP       & 19.4 {\textcolor{red}{\tiny 8.3$\downarrow$}} & 18.3 {\textcolor{red}{\tiny 7.9$\downarrow$}} & 23.5 {\textcolor{red}{\tiny 10.0$\downarrow$}}  & 18.4 {\textcolor{red}{\tiny 7.9$\downarrow$}}   & 25.3 {\textcolor{red}{\tiny 10.9$\downarrow$}}& 17.5 {\textcolor{red}{\tiny 7.5$\downarrow$}} & 15.6 {\textcolor{red}{\tiny 6.7$\downarrow$}} & 19.2 {\textcolor{red}{\tiny 8.2$\downarrow$}}   & 25.3 {\textcolor{red}{\tiny 10.9$\downarrow$}} & 23.4 {\textcolor{red}{\tiny 10.1$\downarrow$}} \\
    \bottomrule
  \end{tabular}
  \label{tab:seg_scene}
  }
\end{table}

\begin{wraptable}{r}{0.5\textwidth}
  \centering
  \scriptsize
  \setlength{\tabcolsep}{8.2pt}
  \caption{Semantic Segmentation Results by Drones' Navigation Strategies Across Different Agent Combinations.}
  \begin{tabular}{l|c|c|c}
    \toprule
    Method & Hover & Patrol & Escort \\
    \midrule
    \multicolumn{4}{l}{\textbf{Vehicle + Infra + Drone}} \\
    \midrule
    When2comm   & 15.1 & 14.8 & 15.8 \\
    CoBEVT      & 34.3 & 40.3 & 28.4 \\
    Where2comm  & 31.7 & 33.8 & 27.6 \\
    V2XViT      & 32.7 & 40.5 & 38.9 \\
    HEAL        & 34.3 & 40.3 & 38.4 \\
    STAMP       & 26.3 & 32.1 & 30.7 \\
    \midrule
    \multicolumn{4}{l}{\textbf{Vehicle + Infra}} \\
    \midrule
    When2comm   & 13.8 {\textcolor{red}{\tiny 1.3$\downarrow$}} & 13.4 {\textcolor{red}{\tiny 1.4$\downarrow$}} & 14.4 {\textcolor{red}{\tiny 1.4$\downarrow$}} \\
    CoBEVT      & 28.2 {\textcolor{red}{\tiny 6.2$\downarrow$}} & 33.0 {\textcolor{red}{\tiny 7.3$\downarrow$}} & 23.3 {\textcolor{red}{\tiny 5.1$\downarrow$}} \\
    Where2comm  & 26.0 {\textcolor{red}{\tiny 5.7$\downarrow$}} & 27.8 {\textcolor{red}{\tiny 6.1$\downarrow$}} & 22.6 {\textcolor{red}{\tiny 5.0$\downarrow$}} \\
    V2XViT      & 26.8 {\textcolor{red}{\tiny 5.9$\downarrow$}} & 33.2 {\textcolor{red}{\tiny 7.3$\downarrow$}} & 31.9 {\textcolor{red}{\tiny 7.0$\downarrow$}} \\
    HEAL        & 28.1 {\textcolor{red}{\tiny 6.2$\downarrow$}} & 33.1 {\textcolor{red}{\tiny 7.3$\downarrow$}} & 31.5 {\textcolor{red}{\tiny 6.9$\downarrow$}} \\
    STAMP       & 21.5 {\textcolor{red}{\tiny 4.7$\downarrow$}} & 26.3 {\textcolor{red}{\tiny 5.8$\downarrow$}} & 25.1 {\textcolor{red}{\tiny 5.6$\downarrow$}} \\
    \midrule
    \multicolumn{4}{l}{\textbf{Vehicle only}} \\
    \midrule
    When2comm   & 12.7 {\textcolor{red}{\tiny 2.4$\downarrow$}} & 12.5 {\textcolor{red}{\tiny 2.4$\downarrow$}} & 13.2 {\textcolor{red}{\tiny 2.5$\downarrow$}} \\
    CoBEVT      & 24.0 {\textcolor{red}{\tiny 10.3$\downarrow$}} & 28.2 {\textcolor{red}{\tiny 12.1$\downarrow$}} & 19.9 {\textcolor{red}{\tiny 8.5$\downarrow$}} \\
    Where2comm  & 22.2 {\textcolor{red}{\tiny 9.5$\downarrow$}} & 23.7 {\textcolor{red}{\tiny 10.1$\downarrow$}} & 19.4 {\textcolor{red}{\tiny 8.3$\downarrow$}} \\
    V2XViT      & 22.9 {\textcolor{red}{\tiny 9.8$\downarrow$}} & 28.3 {\textcolor{red}{\tiny 12.1$\downarrow$}} & 27.2 {\textcolor{red}{\tiny 11.7$\downarrow$}} \\
    HEAL        & 24.1 {\textcolor{red}{\tiny 10.3$\downarrow$}} & 28.2 {\textcolor{red}{\tiny 12.1$\downarrow$}} & 26.9 {\textcolor{red}{\tiny 11.5$\downarrow$}} \\
    STAMP       & 18.4 {\textcolor{red}{\tiny 7.9$\downarrow$}} & 22.5 {\textcolor{red}{\tiny 9.6$\downarrow$}} & 21.4 {\textcolor{red}{\tiny 9.2$\downarrow$}} \\
    \bottomrule
  \end{tabular}
  \label{tab:seg_drone}
\end{wraptable}

Table \ref{tab:seg_scene} presents semantic segmentation performance across various conditions. For semantic segmentation, removing drone agents causes approximately 4.5\%-6.5\% decreases across most models and conditions. When comparing vehicle-only to vehicle+infrastructure, we observe an additional 3.0\$-4.5\$ drop, indicating that while drones provide valuable elevated perspectives for segmentation tasks, infrastructure agents also contribute significantly to boundary delineation and contextual understanding.

Table \ref{tab:seg_drone} reveals that drone flight patterns significantly impact semantic segmentation performance. The patrol pattern, where drones follow predetermined routes, yields the highest baseline performance across all models. Removing drones during patrol scenarios causes substantial performance drops (5.8\%-7.3\%). The vehicle-only configuration experiences the most severe degradation (9.6\%-12.1\%) in patrol scenarios, suggesting that this pattern provides complementary information that cannot be recovered from ground perspectives. For escort patterns, where drones follow specific vehicles, the performance drops are more consistent across models, with 5.0\%-7.0\% decreases when removing drones. The hover pattern shows slightly less sensitivity to drone removal (4.7\%-6.2\%), suggesting that the aerial perspectives are limited in the hover mode because the drones are stationary.

In conclusion, our comprehensive ablation studies demonstrate that each agent type contributes uniquely to perception performance, with their relative importance varying across environmental conditions, scenarios, and perception tasks. Overall, both drone agents and infrastructure agents provide valuable perspectives multi-agent collaborative perception.

\section{Dataset Visualization}\label{sec:dataset_vis}

\begin{figure}[!ht]
    \centering
    \includegraphics[width=\linewidth]{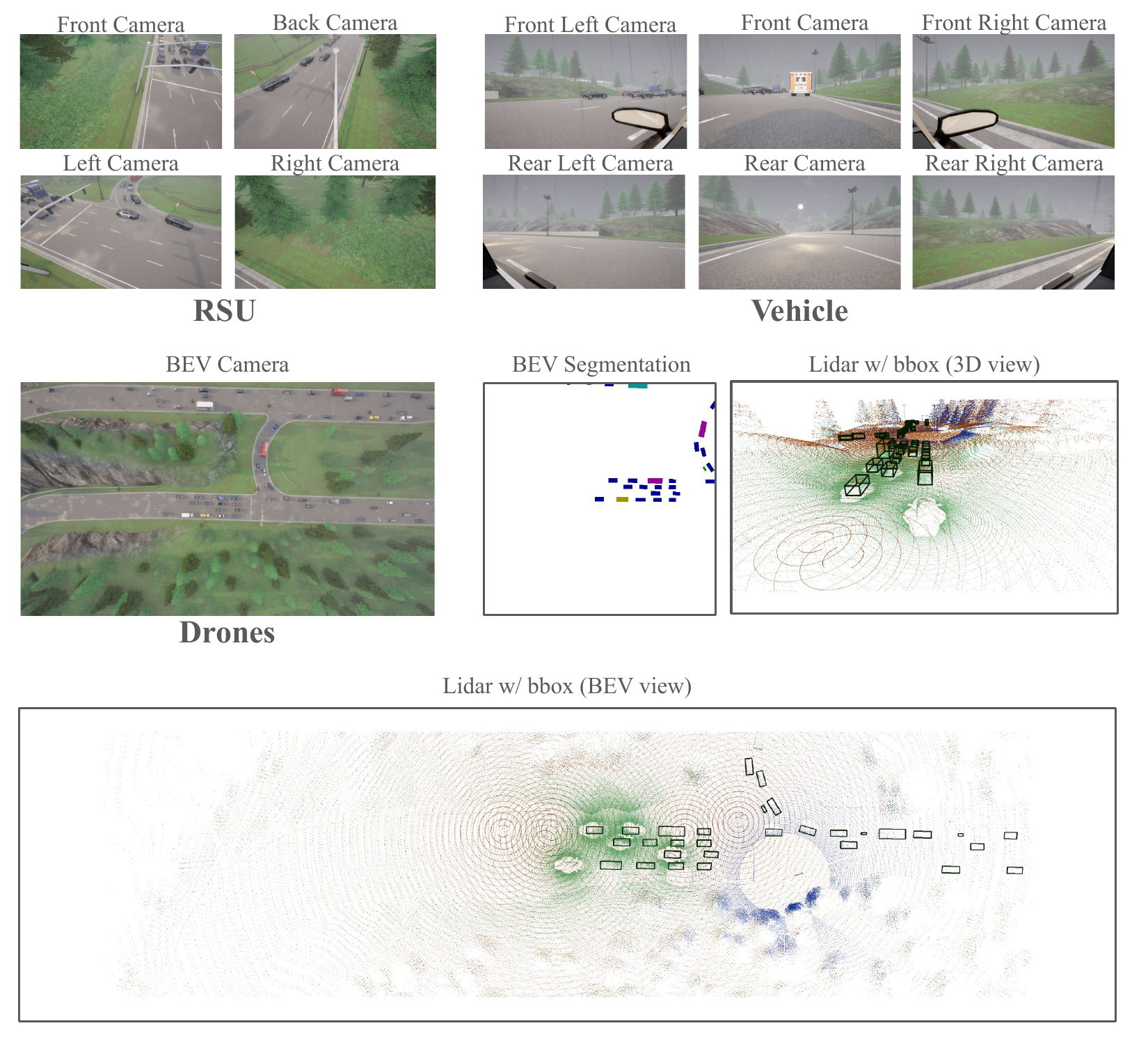}
    \caption{Visualization of some representative data of a single timestamp of the AirV2X-Perception dataset.
    Note that for each agent type from RSU, vehicle, and drone, only one agent is chosen for visualization.}
    \label{fig:data_vis}
\end{figure}

\begin{figure}[!ht]
    \centering
    \includegraphics[width=\linewidth]{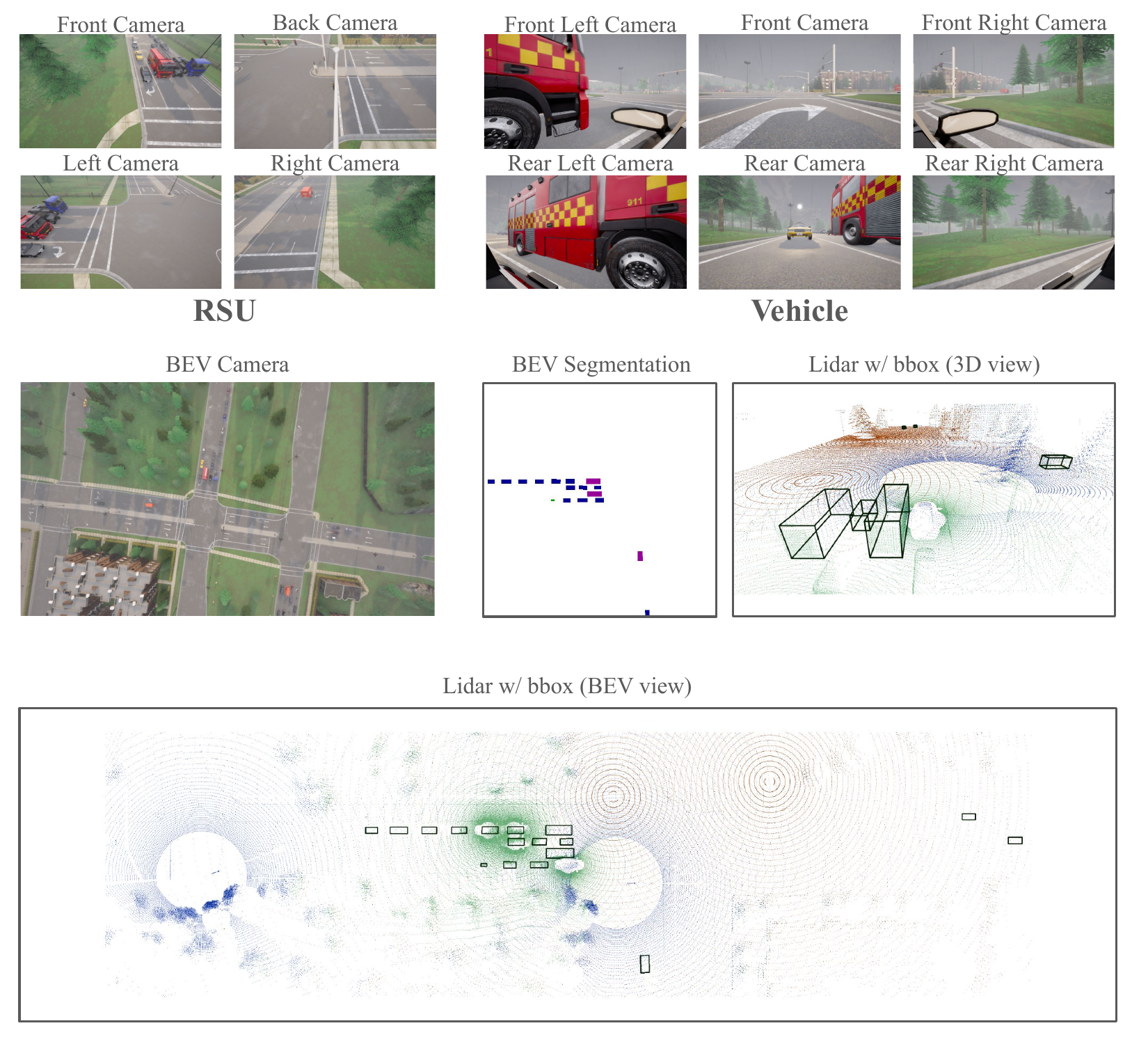}
    \caption{Visualization of some representative data of a single timestamp of the AirV2X-Perception dataset.
    Note that for each agent type from RSU, vehicle, and drone, only one agent is chosen for visualization.}
    \label{fig:data_vis_2}
\end{figure}

Figures~\ref{fig:data_vis} and~\ref{fig:data_vis_2} provide qualitative insight into the breadth and fidelity of the \textit{AirV2X-Perception} dataset. 
Each scene is captured synchronously by heterogeneous sensing platforms—including road-side units (RSUs), connected vehicles, and drones. \emph{(i)}  The top panels in both figures highlight the raw RGB imagery acquired by the surround camera matrix on RSUs and vehicles.  \emph{(ii)}  The middle panels display bird’s-eye-view (BEV) camera image from the drone (left), a semantic BEV map (middle), and a perspective LiDAR point cloud with 3-D bounding boxes (right). \emph{(iii)}  The bottom panel displays the full LiDAR sweep onto the BEV plane axis-aligned bounding boxes.
For the complete dataset, please refer to our open-sourced dataset link\footnote{\url{https://huggingface.co/datasets/xiangbog/AirV2X-Perception}}.


\section{Beyond the dataset}

\subsection{Drone-Assisted Vehicle-to-Everything (V2X) for Autonomous Driving}

The convergence of autonomous driving technologies with unmanned aerial vehicles (UAVs) presents compelling opportunities for intelligent transportation systems. The global drone industry is experiencing remarkable growth, with annual UAV shipments projected to reach 9.5 million units by 2029~\cite{GlobalDrone}, corresponding to a market value of \$35-54 billion in 2024. Within the mobility domain, UAV-assisted logistics and transportation services alone are expected to double from \$5.3 billion in 2019 to \$11 billion by 2026~\cite{afrin2024advancements}. This proliferation of UAV technology is creating new possibilities for drone-assisted Vehicle-to-Everything (V2X) frameworks to enhance autonomous driving capabilities globally.

V2X communication, encompassing interactions between vehicles and their environment (other vehicles, infrastructure, pedestrians, networks), forms the foundation of connected autonomous driving. However, accommodating the growing number of connected devices and data-intensive services in vehicular networks challenges existing infrastructure~\cite{wang2025uav}. Drone-assisted V2X offers a promising solution by integrating aerial drones as dynamic sensor platforms, communication relays, and edge computing nodes. UAVs effectively add a third dimension to V2X networks, enabling more comprehensive coverage and adaptive architectures than static ground infrastructure alone can provide.

\subsection{Current Research Trends and Future Projections}

Research on drone-assisted vehicular networks has accelerated, exploring innovative protocol designs, efficient resource management, and energy-optimized operations for UAV nodes in V2X ecosystems. Current research prototypes demonstrate UAVs functioning as aerial base stations, relays, or cooperative sensing platforms in connected vehicle environments. Field trials show that drones can wirelessly connect isolated vehicle clusters, extend coverage in rural areas, and provide aerial perspectives to detect hazards beyond a vehicle's line-of-sight. Cooperative perception is emerging as a particularly valuable application—drones equipped with cameras or LiDAR can stream data to nearby vehicles, effectively enabling them to "see" around corners or beyond obstructions.

Looking ahead, sixth-generation (6G) wireless architectures are expected to natively support airborne communication nodes, facilitating real-time coordination between dense drone swarms and ground vehicles~\cite{kavas2022v2x}. Researchers anticipate advanced UAV traffic management systems and dynamically reconfigurable airborne base stations that adapt to changing traffic conditions. Advancements in AI are projected to enhance multi-UAV collaboration, enabling autonomous drone swarms to optimize their positioning for network coverage and data collection. The trajectory of research suggests that drone-assisted V2X will evolve from today's experimental implementations to become a fundamental component of smart transportation within the decade.

\subsection{Advantages over Traditional RSUs and Vehicle-Only Systems}

Drone-assisted V2X systems offer advantages over traditional roadside units (RSUs) and purely vehicle-based networks. Their primary strength lies in dynamic adaptability: unlike fixed RSUs, UAVs can be repositioned as needed to provide coverage in response to changing traffic or network conditions. This on-demand deployment reduces the need for ubiquitous physical infrastructure while enabling adaptive network scaling. Studies demonstrate that "flying RSUs" significantly improve connectivity in sparse vehicular networks by filling coverage gaps between distant ground nodes~\cite{hadiwardoyo2019modelling}—particularly valuable in rural areas or developing regions where fixed infrastructure deployment is impractical.

From a sensing perspective, drones provide superior vantage points compared to vehicle-mounted sensors alone. Autonomous vehicles' onboard cameras, radar, and LiDAR have limited range and are vulnerable to occlusions from buildings or large vehicles. UAV-mounted sensors mitigate these limitations by observing the environment from above, seeing over obstacles and surveying broader areas simultaneously. This aerial perspective enables more comprehensive situational awareness when integrated with vehicle data.

Communication performance also improves with drone assistance. Signal propagation for V2X radio is often hindered by buildings, terrain, or dense traffic, especially in urban environments. Aerial relays enjoy clearer line-of-sight paths and can maintain simultaneous links with multiple vehicles from elevated positions. By serving as intermediate nodes, drones reduce the number of hops or transmission distances, thereby lowering latency and increasing data rates.

Finally, drone integration can be cost-effective compared to deploying numerous fixed sensors and RSUs. While individual drones represent sophisticated technology investments, their mobility allows them to cover multiple locations over time and be shared among many users as a service. This reduces the need for permanently installed infrastructure that might be underutilized during off-peak hours. In scenarios like temporary events, construction zones, or disaster response, this agility and efficiency far outperform static, traditional infrastructure approaches.

\subsection{Technical Challenges and Limitations}

Despite their promise, drone-assisted V2X systems face several technical challenges:

\begin{itemize}[leftmargin=8pt, itemsep=2pt, parsep=2pt, topsep=0pt, partopsep=0pt]
\item \textbf{Latency and real-time communication:} Supporting safety-critical autonomous driving applications demands ultra-low latency. Introducing drones as relays adds new sources of delay~\cite{gupta2024latency}. The entire process from capture to broadcast must occur within milliseconds, requiring optimized communication protocols and careful scheduling of V2X message transmissions.

\item \textbf{Energy constraints:} Limited battery life fundamentally restricts most UAVs to 20-40 minutes of flight time, constraining their endurance for continuous V2X support. Frequent battery swaps or recharging would be required for persistent coverage, while energy budgets also limit onboard sensing and computing capabilities. Energy-efficient hardware and operations (including automated docking stations and solar-powered platforms) remain active research areas.

\item \textbf{Safety and airspace conflict:} UAVs must avoid collisions with other aircraft and prevent hazards to people and property below. Mid-air collision avoidance requires reliable detect-and-avoid systems, especially at low altitudes around buildings and traffic. Dedicated UAV-to-UAV communication links have been proposed to coordinate movements and prevent incidents. Robust fail-safe protocols (automatic parachutes, controlled emergency landings) are essential to mitigate risks from battery depletion or malfunction.

\item \textbf{Coordination and scalability:} Managing drone fleets alongside thousands of connected vehicles introduces complex coordination challenges. UAVs must synchronize their trajectories, sensing tasks, and communication resources to maximize coverage without interference. City-wide deployments might require dozens or hundreds of drones, demanding sophisticated aerial traffic management systems. Research continues to explore swarm formation control and adaptive networking algorithms.

\item \textbf{Security and privacy:} Drone integration expands the attack surface of vehicular networks. Communication links between UAVs and vehicles or infrastructure may be vulnerable to eavesdropping, jamming, or spoofing without proper security measures. Additionally, drone-mounted cameras and sensors may capture sensitive data about individuals or businesses, raising privacy concerns that must be addressed through both regulatory frameworks and privacy-preserving technical designs. \end{itemize}

In summary, the integration of unmanned aerial vehicles with vehicle-to-everything communications creates research opportunities driven by both practical needs and technical challenges. While drone-assisted V2X offers compelling advantages in enhanced perception, flexible coverage, and improved communication reliability, it faces substantial hurdles in energy efficiency, latency management, safety, coordination, and security. These challenges establish a rich research landscape spanning communications, sensing, control systems, energy management, and cybersecurity. By addressing these interconnected concerns, researchers can advance drone-assisted V2X from experimental prototypes to practical implementations, ultimately transforming autonomous transportation with dynamic aerial support that overcomes the limitations of traditional ground-based approaches~\citep{wang2025uav, kavas2022v2x}.

\end{document}